\documentclass[sigconf]{acmart}
\usepackage{hyperref}
\usepackage{theoremref}
\usepackage{paralist}
\usepackage{scalerel}

\usepackage[super]{nth}
\usepackage{amsmath}
\usepackage{bm}
\usepackage{array}
\usepackage{makecell}
\usepackage{multirow}
\usepackage{tabularx}
\usepackage{graphicx}
\usepackage{caption}
\usepackage{subcaption}
\usepackage{url}
\usepackage{enumitem}
\usepackage{float}
\usepackage[flushleft]{threeparttable}
\usepackage{tablefootnote}
\usepackage{mathrsfs}

\usepackage[ruled,vlined,linesnumbered]{algorithm2e}
\usepackage{xcolor}

\DeclareMathOperator*{\argmax}{arg\,max}

\usepackage{geometry}
\usepackage{lipsum}    
\usepackage[hang]{footmisc}
\usepackage{lipsum}
\newtheorem{problem}{Problem}

\AtBeginDocument{%
  \providecommand\BibTeX{{%
    \normalfont B\kern-0.5em{\scshape i\kern-0.25em b}\kern-0.8em\TeX}}}

\setcopyright{acmcopyright}
\begin{document}
\setlength{\abovedisplayskip}{3.5pt plus 0pt}%
\setlength{\belowdisplayskip}{3.5pt plus 0pt}%
\title{Imbalanced Graph Classification via \\Graph-of-Graph Neural Networks}
\renewcommand{\shorttitle}{Graph Representation Learning for Imbalanced Graph Classification}
% \title{Imbalanced Graph Classification via \\Graph-of-Graph Neural Networks}


% \author{Anonymous Author(s)}
% \email{Anonymous Email(s)}
% \affiliation{%
%   \institution{Anonymous Affiliation(s)}
%   \country{}
% }

\author{Yu Wang}
\email{yu.wang.1@vanderbilt.edu}
\affiliation{%
  \institution{Vanderbilt University}
  \country{}
}

\author{Yuying Zhao}
\email{yuying.zhao@vanderbilt.edu}
\affiliation{%
  \institution{Vanderbilt University}
  \country{}
}

\author{Neil Shah}
\email{nshah@snap.com}
\affiliation{%
  \institution{Snap Research}
  \country{}
}

\author{Tyler Derr}
\email{tyler.derr@vanderbilt.edu}
\affiliation{%
  \institution{Vanderbilt University}
  \country{}
}

\renewcommand{\shortauthors}{Yu Wang, Yuying Zhao, Neil Shah, \& Tyler Derr}



\begin{abstract}
Graph Neural Networks (GNNs) have achieved unprecedented success in identifying categorical labels of graphs. However, most existing graph classification problems with GNNs follow the protocol of balanced data splitting, which misaligns with many real-world scenarios in which some classes have much fewer labels than others. Directly training GNNs under this imbalanced scenario may lead to uninformative representations of graphs in minority classes, and compromise the overall classification performance, which signifies the importance of developing effective GNNs towards handling imbalanced graph classification. Existing methods are either tailored for non-graph structured data or designed specifically for imbalanced node classification while few focus on imbalanced graph classification. To this end, we introduce a novel framework, Graph-of-Graph Neural Networks (G$^2$GNN), which alleviates the graph imbalance issue by deriving extra supervision globally from neighboring graphs and locally from stochastic augmentations of graphs. Globally, we construct a graph of graphs (GoG) based on kernel similarity and perform GoG propagation to aggregate neighboring graph representations. Locally, we employ topological augmentation via masking node features or dropping edges with self-consistency regularization to generate stochastic augmentations of each graph that improve the model generalibility. Extensive graph classification experiments conducted on seven benchmark datasets demonstrate our proposed G$^2$GNN outperforms numerous baselines by roughly 5\% in both F1-macro and F1-micro scores. Open-source code can be found at \href{https://github.com/YuWVandy/G2GNN}{\textcolor{blue}{https://github.com/YuWVandy/G2GNN}}.
%\vspace{-2ex}
\end{abstract}

%\keywords{\vspace{-1.5ex}graph classification, graph neural network, graph of graphs}
 %imbalance graph classification



%Note: Put here as a placeholder and can be updated


\copyrightyear{2022}
\acmYear{2022}
\setcopyright{acmcopyright}\acmConference[CIKM '22]{Proceedings of the 31st ACM International Conference on Information and Knowledge Management}{October 17--21, 2022}{Atlanta, GA, USA}
\acmBooktitle{Proceedings of the 31st ACM International Conference on Information and Knowledge Management (CIKM '22), October 17--21, 2022, Atlanta, GA, USA}
\acmPrice{15.00}
\acmDOI{10.1145/3511808.3557356}
\acmISBN{978-1-4503-9236-5/22/10}

%   <concept>
%       <concept_id>10002951.10003317</concept_id>
%       <concept_desc>Information systems~Information retrieval</concept_desc>
%       <concept_significance>500</concept_significance>
%       </concept>
%   <concept>
%       <concept_id>10002951.10003227.10003351</concept_id>
%       <concept_desc>Information systems~Data mining</concept_desc>
%       <concept_significance>500</concept_significance>
%       </concept>
%  </ccs2012>

% % \ccsdesc[500]{Information systems~Information retrieval}
% \ccsdesc[500]{Information systems~Data mining}%\vspace{-2.25ex}}

% \keywords{Graph representation learning, graph neural networks, over-smoothing, tree decomposition, multi-hop dependency}


\maketitle

\vspace{-1ex}\section{Introduction}\label{sec-introduction}
Employing graph representations for classification has recently attracted significant attention due to the emergence of Graph Neural Networks (GNNs) associated with its unprecedented power in expressing graph representations~\cite{powerfulgnn}. A typical GNN architecture for graph classification begins with an encoder that extracts node representations by propagating neighborhood information followed by pooling operations that integrate node representations into graph representations, which are then fed into a classifier to predict graph labels~\cite{fairgraph}. Although numerous GNN variants have been proposed by configuring different propagation and pooling schemes, most works are framed under the setting of balanced data-split where an equal number of labeled graphs are provided as the training data for each class~\cite{song2021graph}. However, collecting such balanced data tends to be time-intensive and resource-expensive, and thus are often
impossible in reality~\cite{leevy2018survey}.

In many real-world graph datasets, the distribution of graphs across classes varies from a slight bias to a severe imbalance where a large portion of classes contain a limited number of labeled graphs (minority classes) while few classes contain enough labeled graphs~\cite{gpn, yu2022reconstructing} (majority classes). For example, despite the huge chemical space, few compounds are labeled active with
the potential to interact with a target biomacromolecule; the remaining majority are labeled inactive~\cite{idakwo2020structure, rozemberczki2022chemicalx, liu2022interpretable}. Since most GNNs are designed and evaluated on balanced datasets, directly employing them on imbalanced datasets would compromise the overall classification performance. As one sub-branch of deep learning on graph-structured data, GNNs similarly inherit two severe problems from traditional deep learning on imbalanced datasets: inclination to learning towards majority classes~\cite{johnson2019survey} and poor generalization from given scarce training data to abounding unseen testing data~\cite{song2021graph, zhao2021synergistic}. Aiming at these two challenges, traditional solutions include augmenting data via under- or over-sampling~\cite{chawla2004special, van2007experimental}, assigning weights to adjust the portion of training loss of different classes~\cite{thai2010cost}, and constructing synthetic training data via interpolation over minority instances to balance the training data~\cite{smote}. However, these methods have been primarily designed on point-based data and their performance on graph-structured data is unclear.

Imbalance on graph-structured data could lie either in the node or graph domain where nodes (graphs) in different classes have different amount of training data. Nearly all related GNN works focus on imbalanced node classification by either pre-training or adversarial training to reconstruct the graph topology~\cite{DRGCN, RECT, graphsmote, wang2021distance}, while to the best of our knowledge, imbalanced graph classification with GNNs remains largely unexplored. On one hand, unlike node classification where we can derive extra supervision for minority nodes from their neighborhoods, graphs are individual instances that are isolated from each other and we cannot aggregate information directly from other graphs by propagation. On the other hand, compared with imbalance on regular grid or sequence data (e.g., images or text) where imbalance lies in feature or semantic domain, the imbalance of graph-structured data could also be attributed to the graph topology since unrepresentative topology presented by limited training graphs may ill-define minority classes that hardly generalize to the topology of diverse unseen testing graphs. To address the aforementioned challenges, we present Graph-of-Graph Neural Networks (G$^2$GNN), which consists of two essential components that seamlessly work together to derive supervision globally and locally. %\yu{Globally, we create novel synthetic graph representations by aggregating neighboring graph representations and locally, we augment each graph via masking node features or dropping edges.} 
In summary, the main contributions are as follows:
\vspace{-0.5ex}
\begin{itemize}[leftmargin=*]
    \item \textbf{Problem:}
    We investigate the problem of imbalanced graph classification, which is heavily unexplored in the GNN literature.
    %We investigate a novel problem of imbalanced graph classification with GNNs, \yu{which is heavily unexplored in the GNN literature.}
    % In particular, we emphasize its importance in real-world applications and further provide a formal problem definition. 
    
    \item \textbf{Algorithm:} 
    We propose a novel framework G$^2$GNN for imbalanced graph classification, which derives extra supervision by globally aggregating from neighboring graphs and locally augmenting graphs with self-consistency regularization.
    % \yy{not only augmenting graph topology but also involve features, would it be better just use augmenting graph or something else?}
    %We propose a novel framework G$^2$GNN for imbalanced graph classification, which derives extra supervision by globally aggregating information from neighboring graphs and locally augmenting graph topology with self-consistency regularization.
    
    \item \textbf{Experiments:} 
    We perform extensive experiments on various real-world datasets to corroborate the effectiveness of G$^2$GNN on imbalanced graph classification.
\end{itemize}
 
We define imbalanced graph classification problem in section~\ref{sec-problemform} and related work in section~\ref{sec-relatedwork}. The proposed framework, G$^2$GNN, is given in Section~\ref{sec-framework}, consisting of global graph of graph construction/propagation %module 
and local graph augmentation. % module. 
In Section~\ref{sec: experiment}, we conduct extensive experiments to validate the effectiveness of G$^2$GNN. Finally, we conclude and discuss future work in Section~\ref{sec-conclusion}.
\section{Problem Formulation}\label{sec-problemform}
%In this section, we introduce notations and formulate the problem of imbalanced graph classification.

Let $G = (\mathcal{V}^G, \mathcal{E}^G, \mathbf{X}^G)$ denote an attributed graph with node feature $\mathbf{X}^G\in\mathbb{R}^{|\mathcal{V}^G|\times d}$ and adjacency matrix $\mathbf{A}^{G}\in\mathbb{R}^{|\mathcal{V}^G|\times |\mathcal{V}^G|}$ where $\mathbf{A}^{G}_{ij} = 1$ if there is an edge between nodes $v_i, v_j$ and vice versa. In graph classification, given a set of $N$ graphs $\mathcal{G}=\{G_1, G_2, ..., G_N\}$ with each graph $G_i = (\mathcal{V}^{G_i}, \mathcal{E}^{G_i}, \mathbf{X}^{G_i})$ as defined above and their labels $\mathbf{Y}\in\mathbb{R}^{N\times C}$ where $C$ is the total number of classes, we aim to learn graph representations $\mathbf{P}\in\mathbb{R}^{N\times d'}$ with $\mathbf{P}_i$ for each $G_i \in \mathcal{G}$ that is well-predictive of its one-hot encoded label $\mathbf{Y}_i$. The problem of imbalanced graph classification can be formalized as:
\vspace{-1ex}

\begin{problem}
Given a set of attributed graphs $\mathcal{G}$ with a subset of labeled graphs $\mathcal{G}^\ell$ that are imbalanced among different classes, we aim to learn a graph encoder and classifier $\mathcal{F}: \mathcal{F}(\mathbf{X}^{G_i}, \mathbf{A}^{G_i}) \rightarrow \mathbf{Y}_{i}$ that works well for graphs in both majority and minority classes.
\end{problem}

% \begin{problem}
% Given a set of attributed graphs $\mathcal{G}$ with a subset of $l$ labeled graphs $\mathcal{G}$ that are imbalanced among different classes, we aim to learn a graph encoder and classifier $\mathcal{F}: \mathcal{F}(\mathbf{X}^{G_i}, \mathbf{A}^{G_i}) \rightarrow \mathbf{Y}_{i}$ that works well for graphs in both majority and minority classes.
% \end{problem}

\section{Related Work}\label{sec-relatedwork}
\noindent \textbf{Graph Imbalance Problem.}
Graph imbalance exists in many real-world scenarios~\cite{graphsmote} where graph topology can be harnessed to derive extra supervision for learning graph/node representations. DR-GCN~\cite{DRGCN} handles multi-class imbalance by class-conditional adversarial training and latent distribution regularization. RECT~\cite{RECT} merges a GNN and proximity-based embeddings for the completely-imbalanced setting (i.e., some classes have no labeled nodes during training). GraphSMOTE~\cite{graphsmote} attempts to generate edges by pre-training an edge generator for isolated synthetic nodes generated from SMOTE~\cite{smote}. Most recently, imGAGN~\cite{qu2021imgagn} simulates both distributions of node attributes in minority classes and graph structures via generative adversarial model. However, all of these recent powerful deep learning works are proposed for node imbalance classification. Graph imbalance classification~\cite{pan2013graph}, remains largely unexplored, especially in GNN domain. Therefore, this work tackles this problem and different from previous work, we expect to leverage the graph topology via graph kernels to construct graph of graphs (GoG) and perform propagation on the constructed GoG. % (GoG propagation).

\vspace{1ex}
\noindent\textbf{Graph of Graphs.}
Graphs model entities by nodes and their relationships by edges. Sometimes, nodes at a higher level in a graph can be modeled as graphs at a lower level, which is termed as graph of graphs (GoG)~\cite{non}. This hierarchical relationship was initially used in~\cite{non} to rank nodes in a broader and finer context. Recently, \cite{gu2021modeling} and \cite{wang2020gognn} leverage Graph of Graphs (GoG) to perform link prediction between graphs and graph classification. However, both of them assume the GoG is provided in advance, e.g., \cite{wang2020gognn} constructs edges between two molecule graphs based on their interactions and two drug graphs based on their side effects. Conversely, in this work, we construct a kNN GoG based on graph topological similarity and aggregate neighboring graph information by propagation on the constructed GoG. 
%To the best our knowledge, this is the first work in graph classification constructing GoG from basic graph datasets and leveraging it for propagation.

\vspace{1ex}
\noindent\textbf{Graph Augmentations.}
%\vspace{-1ex}
%\subsection{Data Augmentation.}
% Augmentation aims to expand training data via artificially creating more reasonable virtual data from existing limited data~\cite{mayer2018makes}. Training model with augmented data would increase the generability of the model such that the performance on unseen testing data could also be improved.
Recent years have witnessed successful applications of data augmentation in computer vision (CV)~\cite{shorten2019survey} and natural language processing (NLP)~\cite{feng2021survey}. As its derivative in graph domain, graph augmentation enriches the training %graph 
data~\cite{zhao2021data, zhao2022graph, ding2022data} and therefore can be naturally leveraged to alleviate class imbalance. In this work, we augment graphs by randomly removing edges and masking node features~\cite{graphCL, wang2022graph} to enhance the model generalizability and further employ self-consistency regularization to enforce the model to output low-entropy predictions~\cite{berthelot2019mixmatch}.

% and constrain that the classifier's decision boundary should not pass high-density regions~\cite{berthelot2019mixmatch}.
% \vspace{-2ex}
\section{The Proposed Framework}\label{sec-framework}
\begin{figure}[t]
     \centering
     %\vskip -1.5ex
     \hspace{1.5ex}
    \includegraphics[width=1.05\columnwidth]{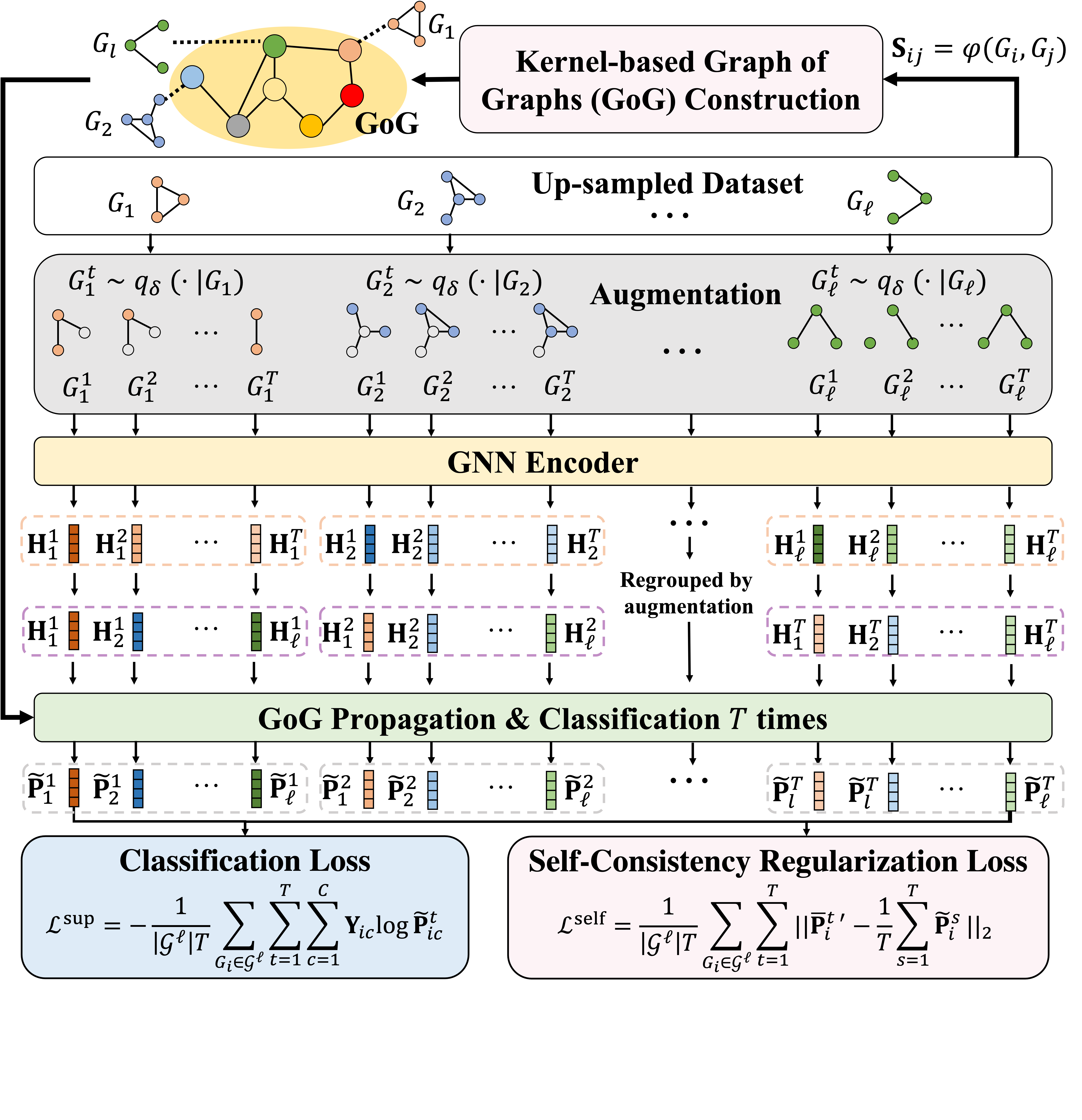}
      \vskip -10ex
      \caption{An overview of the Graph-of-Graph Neural Network  (G$^2$GNN). To reduce imbalance effect on graph classification, we up-sample minority graphs, augment each graph $T$ times followed by a GNN encoder to get their representations and regroup them according to their augmentation order, perform GoG propagation on constructed GoG $T$ times with each time using all graph representations from that specific augmentation $t$, and finally forward the propagated representations through a classifier to compute classification loss and self-consistency regularization loss.}
     %\caption{An overview of the Graph-of-Graph Neural Network  (G$^2$GNN). Here we up-sample minority graphs to reduce imbalance effect, augment each graph $T$ times followed by a GNN encoder to get their representations and regroup them according to their augmentation order, perform GoG propagation on constructed GoG $T$ times with each time using all graph representations from that specific augmentation $t$, and finally forward the propagated representations through classifier to compute classification loss and self-consistency regularization loss.}
     
     \label{fig-framework}
     \vskip -3ex
\end{figure}

 In this section, we introduce our proposed G$^2$GNN framework. Figure~\ref{fig-framework} presents an overview of G$^2$GNN, which is composed of two modules from global and local perspective. Globally, a graph kernel-based GoG construction is proposed to establish a $k$-nearest neighbor (kNN) graph and hence enable two-level propagation, where graph representations are first obtained 
%  by pooling after node-level propagation 
 via a GNN encoder and then neighboring graph representations are aggregated together through the GoG propagation on the established kNN GoG. Locally, we employ graph augmentation via masking node features or removing edges with self-consistency regularization to create novel supervision from stochastic augmentations of each individual graph. The GoG propagation serves as a global governance to retain the model discriminability by smoothing intra-class graphs while separating inter-class graphs. Meanwhile the topological augmentation behaves as a local explorer to enhance the model generalibility in discerning unseen non-training graphs. Next, we introduce details of each module.

%\vspace{-1ex}
%\subsection{Global \td{Imbalance Mitigation}: GoG Construction and Propagation}
\subsection{\hspace{-1.5ex} { Global Imbalance Mitigation: Graph-of-Graph Construction/Propagation}}

% \subsection{\hspace{-1ex}Local Imbalance Mitigation: Self-consistency \\\hspace{-1ex} Regularization via Graph 
% Augmentation}\label{subsec-graphaug}
% \td{just considering ways to perhaps link this notion of global and local towards the goal of solving imbalance problem}
Graph representations obtained by solely forwarding each graph through GNN encoders cannot be well-learned given scarce labeled training graphs in minority classes. Therefore, we construct a GoG to connect independent graphs and perform GoG propagation to aggregate the information of neighboring graphs. The intuition is that feature propagation and aggregation would mimic the way of SMOTE~\cite{smote} and mixup~\cite{mix}, which are two of the most fundamental approaches handling the issue of class imbalance and poor generalizability. Aggregating representations of graphs of the same/different class/classes would simulate the interpolation of SMOTE/mixup with coefficients being determined by the specific graph convolution we use in propagation. In the following, we first introduce the basic GNN encoder to obtain graph representations, which will be later used for GoG propagation. Then we construct GoG and empirically demonstrate its high homophily, which naturally motivates the GoG propagation.

\subsubsection{Basic GNN Encoder}\label{sec-gnnencoder}
% Our graph classifier begins with a graph encoder that utilizes graph topology and node features to learn graph representation. 
In this work, we employ graph isomorphism network (GIN) as the encoder to learn graph representation given
%due to 
its distinguished discriminative power of different topology~\cite{powerfulgnn}. However, our framework holds %straightforwardly 
for any other GNN encoder. One GIN layer is defined as:
\begin{equation}\label{eq:GIN}
    \mathbf{X}^{G_i, l + 1} = \text{MLP}^l((\mathbf{A}^{G_i} + (1 + \epsilon)\mathbf{I})\mathbf{X}^{G_i, l}), \forall l\in\{1,2,..., L\}
\end{equation}
where $\mathbf{X}^{G_i, l}$ is the intermediate node representation at layer $l$, $\mathbf{X}^{G_i, 0} = \mathbf{X}^{G_i}$ is the initial node feature in the graph $G_i$, and MLP is a multi-layer perceptron at layer $l$. After $L$ GIN convolutions, each node aggregates information from its neighborhoods up to $L$ hops away and a readout function integrates node representations into the graph representation $\mathbf{H}_i$ for each graph $G_i$ as:
\begin{equation}\label{eq:pool}
    \mathbf{H}_i = \mbox{READOUT}(\{\mathbf{X}^{G_i, L}_{j}|v_j\in\mathcal{V}^{G_i}\})
\end{equation}
Then we construct a kNN graph on top of each individual graph and perform GoG propagation to borrow neighboring graphs' information. Here we employ global-sum pooling as our READOUT function, which adds all nodes' representations to obtain the graph representation.
% As justified before, computing classification loss with the representation of each individual graph $\mathbf{H}_i$ would compromise the performance on minority graphs. Therefore, we construct a kNN graph on top of each individual graph and perform GoG propagation to borrow neighboring graphs' information.

% \vspace{-1.5ex}
\subsubsection{Graph of Graphs Construction}\label{sec-knnconstruct}
\begin{figure}[t]
     \centering
    %  \vskip -1.5ex
     \hspace{-2ex}
     %\vstretch{1}{\includegraphics[width=.365\textwidth]{figure/kernelhomophily.pdf}}
     \vstretch{1}{\includegraphics[width=.37\textwidth]{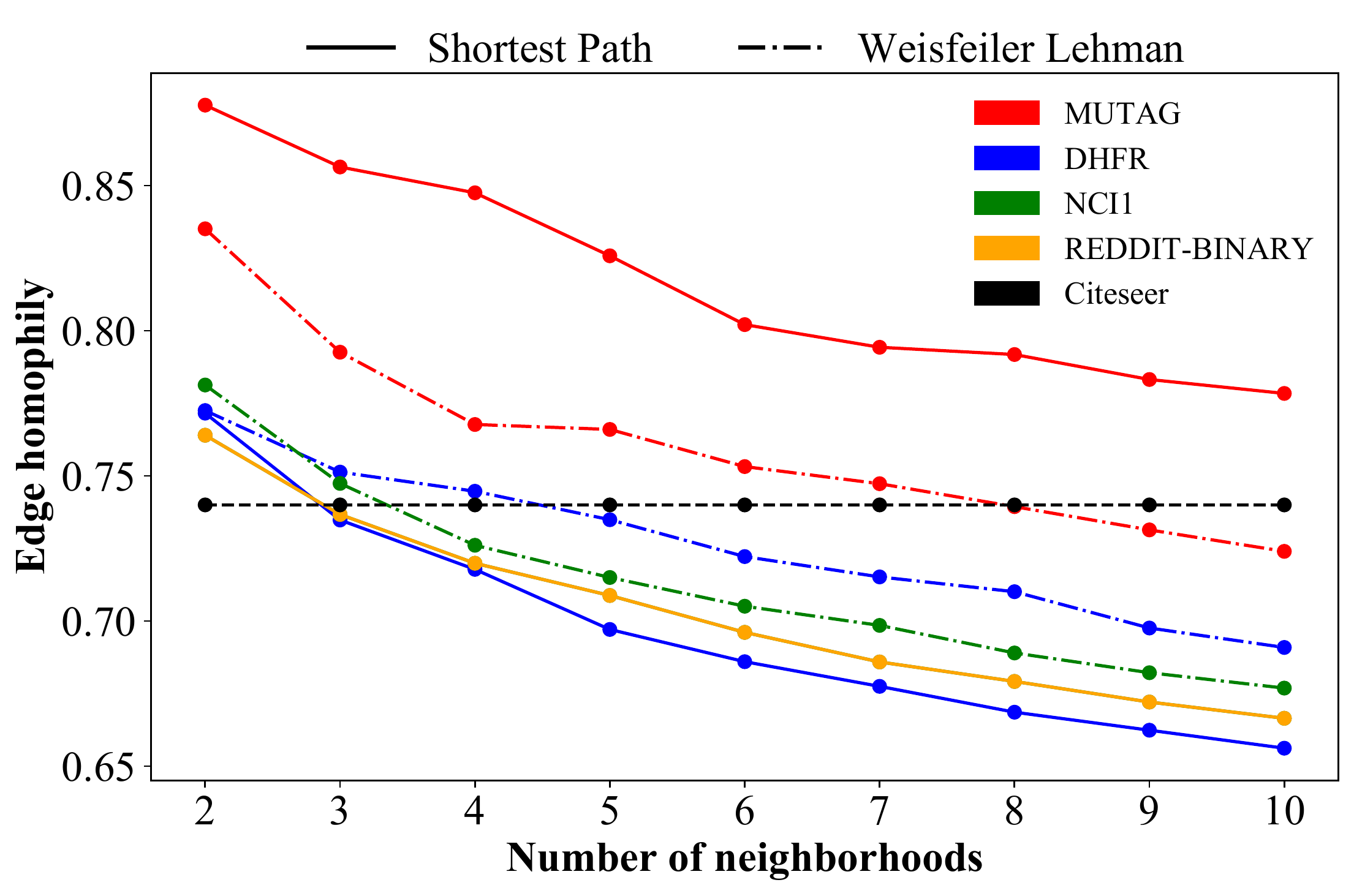}}
     \vskip -2.5ex
     \caption{Edge homophily of constructed kNN GoGs.}% on each dataset.} %graphs.}
     \vskip -2ex
     \label{fig-homoknn}
\end{figure}

Given a set of graphs $\mathcal{G}$, we expect to construct a high-level graph where every graph $G_i \in \mathcal{G}$ is represented by a node and two graphs are linked by an edge if they are similar. In this work, we determine the graph similarity based on their topological similarity since graphs with similar topology typically possess similar functions or belong to the same class such as scaffold hopping~\cite{zheng2021deep} and enzyme identification~\cite{vishwanathan2010graph}. Here we leverage the graph kernel to quantify topological similarity between pairs of graphs~\cite{yanardag2015deep} and further use it to construct GoG. Denote the similarity matrix as $\mathbf{S}\in\mathbb{R}^{N\times N}$ where each entry $\mathbf{S}_{ij}$ measures the topological similarity between each pair of graphs $(G_i, G_j)$ and is computed by the kernel function $\phi$ as:
\begin{equation}\label{eq:kernel}
    \mathbf{S}_{ij} = \phi(G_i, G_j),
\end{equation}
where multiple choices of the kernel function $\phi$ could be adopted here depending on specific types of topological similarity required by downstream tasks and in this work, we choose the Shortest Path Kernel due to its simplicity and effectiveness as demonstrated in Section~\ref{sec: experiment}. Then we construct a kNN graph $\mathcal{G}^{\text{kNN}}$ by connecting each graph $G_i$ with its top-$k$ similar graphs based on the similarity matrix $\mathbf{S}$ and then measure its edge homophily as:
\begin{equation}
    \chi^{\mathcal{G}^{\text{kNN}}} = \frac{|\{(G_i, G_j)\in\mathcal{E}^{\mathcal{G}^{\text{kNN}}}: \mathbf{Y}_i = \mathbf{Y}_j\}|}{|\mathcal{E}^{\mathcal{G}^{\text{kNN}}}|},
\end{equation}
where high $\chi^{\mathcal{G}^{\text{kNN}}}$ means most edges connect graphs of the same class and by varying $k$, we end up with multiple $\mathcal{G}^{\text{kNN}}$ with different homophily level. Figure~\ref{fig-homoknn} visualizes the homophily of $\mathcal{G}^{\text{kNN}}$ constructed using Shortest-Path and Weisfeiler-Lehman kernels on three graph datasets populating in the literature~\cite{graphCL, Infograph}. We can clearly see that edge homophily decreases as $k$ increases because graphs with lower topological similarity have higher chance to be selected as neighborhoods while they likely belong to different classes from corresponding center graphs. However, edge homophily even when $k$ is up to 5 is still in $\lbrack0.7,0.8\rbrack$ and %, which is 
comparable to Citeseer dataset%\footnote{Citeseer is a well-known node classification dataset and commonly used for benchmarking GNNs~\cite{song2021graph}}, 
\footnote{Citeseer: a well-known GNN node classification benchmark dataset~\cite{song2021graph}.}, 
which indicates that most edges in the constructed $\mathcal{G}^{\text{kNN}}$ connects graphs of the same class. Motivated by this observation, we perform GoG propagation on the generated kNN graph $\mathcal{G}^{\text{kNN}}$ to aggregate neighboring graph information.

%\vspace{-1.5ex}
\subsubsection{Graph of Graphs Propagation}\label{subsec-GoGprop}
Denoting the adjacency matrix with added self-loops of the constructed graph $\mathcal{G}^{\text{kNN}}$ as $\hat{\mathbf{A}}^{\text{kNN}} = \mathbf{A}^{\text{kNN}} + \mathbf{I}$ and the corresponding degree matrix as $\hat{\mathbf{D}}^{\text{kNN}}$, the $l^{\text{th}}$-layer GoG propagation is formulated as:
\begin{equation}\label{eq:smoothing}
    \mathbf{P}^{l+1} = (\hat{\mathbf{D}}^{\text{kNN}})^{-1}\hat{\mathbf{A}}^{\text{kNN}}\mathbf{P}^{l}, l\in\{1,2,..., L\}
\end{equation}
where $\mathbf{P}^{0} = \mathbf{H}$ includes representations of all individual graphs $\mathbf{H}_i$ that are previously obtained from GIN followed by the graph pooling, as Eqs.~\eqref{eq:GIN}-\eqref{eq:pool}. Note that here we do not differentiate between layers $l, L$ used in GoG propagation here and layers used in GIN convolution in Section~\ref{sec-gnnencoder} since their difference is straightforward based on the context. After $L$ layers propagation, the representation of a specific graph $\mathbf{P}^{L}_i$ aggregates information from neighboring graphs up to $L$ hops away, which naturally smooths neighboring graphs and their labels by the following theorem~\cite{wang2020unifying}:

% The intuition behind the GoG propagation is smoothing among neighboring graphs since the representation of every graph $\mathbf{P}_{i}^{l+1}$ is the weighted average of its neighboring graphs' representations at previous layer $l$:
% \vspace{-.5ex}
% \begin{equation}\label{eq:propagation}
%     \mathbf{P}_{i}^{l+1} = \frac{1}{\hat{\mathbf{d}_i}}\sum_{G_j\in\hat{\mathcal{N}}_i}\mathbf{P}_j^l,
% \end{equation}
% where $\hat{\mathcal{N}}_i = \{\mathcal{N}_i, G_i\}$ is the set of neighboring graphs of $G_i$ including itself and $\hat{\mathbf{d}_i} = \hat{\mathbf{D}}^{\text{kNN}}_{ii}$ is the degree of graph $G_i$ in the constructed $\mathcal{G}^{\text{kNN}}$. 

% Feature smoothing naturally connects with label smoothing by the following theorem:

\begin{theorem} \thlabel{smoothing}
Suppose that the latent ground-truth mapping $\mathcal{M}: \mathbf{P}^{l}_i \rightarrow \mathbf{Y}_i$ from graph representations to graph labels is differentiable and satisfies $\mu-$Lipschitz constraints, i.e., $|\mathcal{M}(\mathbf{P}_i^l) - \mathcal{M}(\mathbf{P}_j^l)|\le \mu||\mathbf{P}_i^l - \mathbf{P}_j^l||_2$ for any pair of graphs $G_i, G_j$ ($\mu$ is a constant), then the label smoothing is upper bounded by the feature smoothing among graph $G_i$ and its neighboring graphs $\hat{\mathcal{N}}_i$ through~\eqref{error} with an error $\boldsymbol{\epsilon}_i^{l} = \mathbf{P}_i^{l + 1} - \mathbf{P}_i^{l}$:
\end{theorem}
\vspace{-1.5ex}
\begin{equation}\label{error}
   (\underbrace{\hat{\mathbf{d}_{i}}^{-1}\sum_{G_j\in\hat{\mathcal{N}}_i}{\mathbf{Y}_j} - \mathbf{Y}_i}_{\text{Label smoothing}}) - (\underbrace{\hat{\mathbf{d}}_i^{-1}\sum_{G_j\in\hat{\mathcal{N}}_i}o(||\mathbf{P}_j^l - \mathbf{P}_i^l||_2)}_{\text{Feature smoothing}}) \le \mu\boldsymbol{\epsilon}_i^l.
\end{equation}
Proof of Theorem~\ref{smoothing} is provided in~\cite{wang2020unifying}. Specifically, $\boldsymbol{\epsilon}_i^{l}$ quantifies the difference of the graph $G_i$'s representation between ${l}^{\text{th}}$ and ${(l+1)}^{\text{th}}$ propagation, which decreases as propagation proceeds~\cite{li2018deeper} and eventually converges after infinite propagation $\lim_{l\to\infty} \boldsymbol{\epsilon}_i^{l} = 0$~\cite{liu2020towards}. Treating each graph $G_i$ as a node in $\mathcal{G}^{\text{kNN}}$ and its representation $\mathbf{P}_i^l$ gradually converges since $\lim_{l\to\infty} \epsilon_i^{l} = 0$. Such feature smoothing further leads to the label smoothing based on \thref{smoothing}. Therefore propagating features according to \eqref{eq:smoothing} is equivalent to propagating labels among neighboring graphs, which derives extra information for imbalance classification. Given the high-homophily of the $\mathcal{G}^{\text{kNN}}$ in Figure~\ref{fig-homoknn}, i.e, neighboring graphs tend to share the same class, the extra information derived from feature propagation (label propagation) is very likely beneficial to the performance of downstream classification.

\vspace{-1ex}
%\subsection{\hspace{-1ex}Local \td{Imbalance Mitigation}: Graph Augmentation and Self-consistency Regularization}\label{subsec-graphaug}
\subsection{\hspace{-1ex}Local Imbalance Mitigation: Self-consistency \\\hspace{-1ex} Regularization via Graph 
Augmentation}\label{subsec-graphaug}
Even though feature propagation globally derives extra label information for graphs in minority classes from their neighboring graphs, training with limited graph instances still restricts the power of the model in recognizing numerous unseen non-training graphs. To retain the model generalibility, we further leverage two types of augmenting schemes, removing edges and masking node features~\cite{graphCL}, which are respectively introduced in the next.

\subsubsection{Removing Edges:} For each graph $G_i\in\mathcal{G}$, we randomly remove a subset of edges $\widehat{\mathcal{E}}^{G_i}$ from the original edge set $\mathcal{E}^{G_i}$ with probability:
$P(e_{uv}\in\widehat{\mathcal{E}}^{G_i}) = 1 - \delta_{uv}^{G_i}$, where $\delta_{uv}^{G_i}$ could be uniform or adaptive for different edges.  Since uniformly removing edges (i.e., $\delta_{uv}^{G_i} = \delta$) already enjoys a boost over baselines as shown in Section~\ref{subsec-pc}, we leave the adaptive one as future work.

\subsubsection{Masking Node Features:} Instead of directly removing nodes that may disconnect the original graph into several components, we retain the graph structure by simply zeroing entire features of some nodes following~\cite{GRAND, graphCL}. Randomly masking entire features of some nodes enables each node to only aggregate information from a random subset of its neighborhoods multi-hops away, which reduces its dependency on particular neighborhoods. Compared with partially zeroing some feature channels, we empirically find that zeroing entire features generates more stochastic augmentations and achieves better performance. Formally, we randomly sample a binary mask $\eta_j^{G_i}\sim\text{Bernoulli}(1 - \delta_j^{G_i})$ for each node $v_j$ in graph $G_i$ and multiply it with the node feature, i.e., $\widehat{\mathbf{X}}_j^{G_i} = \eta_j^{G_i}\mathbf{X}_j^{G_i}$~\cite{zhao2021synergistic}.
%Same as removing edges, we only consider uniformly masking node features in this work. 
%Collectively, we unify the probability of removing edges $\delta_{uv}^{G_i}$ and the ratio of masking node features $\delta_j^{G_i}$ as augmentation ratio $\delta$ for model simplicity.

For model simplicity, we unify the probability of removing edges and masking node features as a single augmentation ratio $\delta$.
Note that by using these augmentations after feature propagation, features of each node are stochastically mixed with its neighborhoods and create multiple augmented representations, which significantly increases the model generalibility if these augmented representations overlap with unseen non-training data. However, arbitrary modification of graph topology without any regularization could unintentionally introduce invalid or even abnormal topology. Therefore, we leverage self-consistency regularization to enforce the model to output low-entropy predictions~\cite{GRAND}.

\vspace{-1ex}
\subsubsection{Self-Consistency Regularization}
Formally, given a set of $T$ augmented variants of a graph $G_i$, $\widehat{\mathcal{G}}_i = \{G_i^1, G_i^2, ..., G_i^T|G_i^t\sim q_{\delta}(\cdot|G_i)\}$ where $q_{\delta}(\cdot|G_i)$ is the augmentation distribution conditioned on the original graph $G_i$ parameterized by the augmentation ratio $\delta$, we feed them through a graph encoder by Eq.~\eqref{eq:GIN}-\eqref{eq:pool} and the GoG propagation by Eq.~\eqref{eq:smoothing} to obtain their representations $\{\mathbf{P}^{1}_i, \mathbf{P}^{2}_i, ..., \mathbf{P}^{T}_i\}$. More specifically, we forward the set of representations of all $t^{\text{th}}$-augmented graphs $\{\mathbf{H}_i^t|i\in\{1, 2, ..., |\mathcal{G}^{\mathscr{l}}|\}\}$ through GoG propagation parallelly $T$ times to obtain their representations $\{\mathbf{P}_i^t|i\in\{1, 2, ..., |\mathcal{G}^{\mathscr{l}}|\}, t\in\{1, 2, ..., T\}\}$. Then we further apply the classifier to obtain their predicted label distributions $\{\widetilde{\mathbf{P}}_i^t = \sigma(g_{\boldsymbol{\theta}_g}(\mathbf{P}_i^t))|i\in\{1, 2, ..., |\mathcal{G}^{\mathscr{l}}|\}, t\in\{1, 2, ..., T\}\}$ where $\sigma$ is the softmax normalization and $g_{\boldsymbol{\theta}_g}$ is a trainable classifier parametrized by $\boldsymbol{\theta}_g$. After that, we propose to optimize the consistency of predictions among $T$ augmentations for each graph. We first calculate the center of label distribution by taking the average of predicted distribution of all augmented variants for each specific graph $G_i$, i.e., $\hat{\mathbf{P}}_{i} = \frac{1}{T}\sum_{t = 1}^{T}\widetilde{\mathbf{P}}^t_i$. Then we sharpen~\cite{berthelot2019mixmatch} this label distribution center:
\begin{equation}\label{eqself}
    \bar{\mathbf{P}}_{ij} = (\hat{\mathbf{P}}_{ij})^{\tau}/\sum_{c = 1}^{C}(\hat{\mathbf{P}}_{ic})^{\tau}, \forall j\in\{1, 2, ..., C\}, i\in\{1, 2, ..., |\mathcal{G}^{\mathscr{l}}|\}
\end{equation}
where $\tau\in[0,1]$ acts as the temperature to control the sharpness of the predicted label distribution and as $\tau \rightarrow \infty$, the sharpened label distribution of each graph approaches a one-hot distribution and hence becomes more informative. Then the self-consistency regularization loss for the graph $G_i$ is formulated as the average $L_2$ distance between the predicted distribution of each augmented graph $\widetilde{\mathbf{P}}^{t}_i$ and their sharpened average predicted distribution:
\begin{equation}\label{eqself}
    \mathcal{L}^{\text{self}}_{i} = \frac{1}{T}\sum_{t = 1}^{T}||\bar{\mathbf{P}}_i - \widetilde{\mathbf{P}}_i^t||_2.
\end{equation}

Optimizing~\eqref{eqself} requires the encoder and classifier to output similar predicted class distribution of different augmentations of each graph to the center one; this prevents the decision boundary of the whole model from passing through high-density regions of the marginal data distribution~\cite{grandvalet2004semi}. %Furthermore, 
Also, as we increase $\tau$, we can enforce the model to output low-entropy (high-confidence) predictions.

\vspace{-1ex}
\subsection{Objective Function and Prediction}\label{subsec-obj}
The overall objective function of G$^2$GNN is formally defined as:
\vspace{-1ex}
\begin{equation}\label{eq: obj}
\footnotesize
     \mathcal{L} =\underbrace{-\frac{1}{|\mathcal{G}^{\mathscr{l}}|T}{\sum_{G_i\in \mathcal{G}^{\mathscr{l}}}}\sum_{t = 1}^{T}\sum_{c = 1}^{C}{\mathbf{Y}_{ic}\log\widetilde{\mathbf{P}}^{t}_{ic}}}_{\mathcal{L}^{\text{sup}}} + \underbrace{\frac{1}{|\mathcal{G}^{\mathscr{l}}|T}\sum_{G_i\in \mathcal{G}^{\mathscr{l}}}\sum_{t = 1}^{T}||\bar{\mathbf{P}}_i - \widetilde{\mathbf{P}}_i^t||_2}_{\mathcal{L}^{\text{self}}},
\end{equation}
where $\mathcal{L}^{\text{sup}}$ is the cross entropy loss over all training graphs in $\mathcal{G}^{\mathscr{l}}$ with known label information as previously defined  with $C$ graph classes to be predicted, and  $\mathcal{L}^{\text{self}}$ is the self-consistency regularization loss defined by Eq.~\eqref{eqself} over all training graphs. 

To predict classes of graphs in validation/testing set, instead of forwarding each individual unlabeled graph through the already-trained encoder $f_{\boldsymbol{\theta}_f}$ and the classifier $g_{\boldsymbol{\theta}_g}$ to predict its label, we first generate $T$ augmented variants of each unlabeled graph $\widehat{\mathcal{G}}_i=\{G_i^1, G_i^2, ..., G_i^T|G_i^t \sim q_{\delta}(\cdot|G_i)\}, \forall G_i\in\mathcal{G}/\mathcal{G}^{\mathscr{l}}$ following Section~\ref{subsec-graphaug} and then collectively forward the group of augmented graphs through $f_{\boldsymbol{\theta}_f}$, GoG propagation and the classifier $g_{\boldsymbol{\theta}_g}$ to obtain their predicted label distribution $\{\widetilde{\mathbf{P}}_i^1, \widetilde{\mathbf{P}}_i^2, ...., \widetilde{\mathbf{P}}_i^T\}$, then the final predicted distribution of graph $\mathcal{G}_i$ is averaged over all augmented variants as $\frac{1}{T}\sum_{t = 1}^{T}\widetilde{\mathbf{P}_i}^t, \forall \mathcal{G}_i \in \mathcal{G}/\mathcal{G}^{\mathscr{l}}$ and the final predicted class is the one that owns the highest class probability, i.e., $y_i = \argmax\limits_{j\in\{1, 2, ..., C\}}\frac{1}{T}\sum_{t = 1}^{T}\widetilde{\mathbf{P}}_{ij}^t$.

\vspace{-1.5ex}
\subsection{Algorithm}
In Algorithm~\ref{alg-GNNGG}, we present a holistic overview of the key stages in the proposed G$^2$GNN framework. Note that the GoG propagation and the graph augmentation with self-consistency regularization are both proposed to create more supervision from scarce minority training graphs, which can only handle the poor generalization problem. To avoid the problem of inclination to learning towards majority classes as mentioned in Section~\ref{sec-introduction}, we up-sample minority labeled graphs till the graphs in training and validation set are both balanced among different classes before starting the whole training processes as step 3 shows here. Balancing the labeled graphs in training set cannot only balance the training loss computed by Eq.~\eqref{eq: obj} but also provide sufficient graphs from minority class to construct GoG. Otherwise given only few graphs in the minority class, the top-$k$ similar graphs to one graph in minority class would be more likely come from majority class, which would further cause inter-class feature smoothing when performing GoG propagation and hence compromise the classification performance. Balancing the labeled graphs in validation set could avoid the imbalanced bias introduced in determining which model should be preserved for later evaluation. Note that in Table~\ref{table-result}, we show that even equipping other baselines with up-sampling to remove the imbalanced training bias, G$^2$GNN still achieves better performance, which demonstrates that the performance improvement is not solely caused by the technique of up-sampling but also by the proposed GoG propagation and augmentation with self-consistency regularization.

\setlength{\textfloatsep}{4pt}
\begin{algorithm}[t]
 \DontPrintSemicolon
 \footnotesize
 \KwIn{The imbalanced set of labeled graphs $\mathcal{G}^{\mathscr{l}}$, the kernel function $\phi$, the augmentation distribution $q_{\delta}$, the encoder $f_{\boldsymbol{\theta}_f}$ and the classifier $g_{\boldsymbol{\theta}_g}$ with their learning rate $\alpha_f, \alpha_g$.}
 
%  \KwOut{$f_{\boldsymbol{\theta}_f}$ and $g_{\boldsymbol{\theta}_g}$ with learned parameters $\boldsymbol{\theta}_f, \boldsymbol{\theta}_g$}

%  Initialize the model parameters $\boldsymbol{\theta}_f, \boldsymbol{\theta}_g$\;

 Compute pairwise similarity matrix $\mathbf{S}$ by Eq.~\eqref{eq:kernel} and construct $\mathcal{G}^{\text{kNN}}$ following Section~\ref{sec-knnconstruct}\;   
 
 Up-sample minority graphs in $\mathcal{G}^{\mathscr{l}}$ for both training and validation sets\;
%  \SetAlgoCaptionSeparator{}
 \While{not converged}
 {
 
    \For{mini-batch of graphs $\mathcal{G}^{B} = \{G_i|G_i\sim\mathcal{G}^{\mathscr{l}}, i = \{1,2,..., |\mathcal{G}^{B}|\}\}$}
    {
        Find top-$k$ similar graphs for each $G_i\in\mathcal{G}^B$ based on $\mathbf{S}$ and incorporate them into $\mathcal{G}^{B}$\tcp*{Section~\ref{sec-knnconstruct}}

        Obtain the subgraph $\mathcal{G}^{\text{kNN}, B}$ from $\mathcal{G}^{\text{kNN}}$ induced by graphs in $\mathcal{G}^{B}$\;
    
        For each $G_i\in\mathcal{G}^B$, generate $T$ augmented graphs $\widehat{\mathcal{G}}_i=\{G_i^1, G_i^2, ..., G_i^T|G_i^t\sim q_{\delta}(\cdot|G_i)\}$ \tcp*{Section~\ref{subsec-graphaug}}
    
        Apply graph encoder $f_{\boldsymbol{\theta}_f}$ by Eqs.~\eqref{eq:GIN}-\eqref{eq:pool}, the GoG propagation by Eq.~\eqref{eq:smoothing}, and the classifier $g_{\boldsymbol{\theta}_g}$ to predict graph class distribution $\{\widetilde{\mathbf{P}}^{t}_i|G_i\in\mathcal{G}^{\mathscr{l}}, t\in T\}$\tcp*{Section~\ref{subsec-GoGprop}}
        
        Compute loss by Eq.~\eqref{eq: obj} and update parameters 
        
        $\boldsymbol{\theta_g} \leftarrow \boldsymbol{\theta_g}-
        \alpha_g*\nabla_{\boldsymbol{\theta}_g}\mathcal{L}$, $\boldsymbol{\theta_f} \leftarrow \boldsymbol{\theta_f}-
        \alpha_f*\nabla_{\boldsymbol{\theta}_f}\mathcal{L}$\tcp*{Section~\ref{subsec-obj}}
    }
 }
\caption{\small The algorithm of G$^2$GNN}
\label{alg-GNNGG}
\end{algorithm}

\vspace{-4ex}
% \section{Complexity Analysis}
\subsection{Complexity Analysis}
Next, we compare our proposed G$^2$GNN with vanilla GNN-based encoders by analyzing the time and model compelxity. Since we employ shortest path kernel for all experiments in this work, we only analyze our models with this specific graph kernel. %variant.

In comparison to vanilla GNN-based encoders, additional computational requirements come from three components: kernel-based GoG construction and topological augmentation. In kernel-based GoG construction, applying shortest path kernel to calculate the similarity between every pair of graphs requires $O(n^3)$~\cite{borgwardt2005shortest} time and thus the total time complexity of this part is $O({|\mathcal{G}| \choose 2}\tilde{n}^3)$ ($\tilde{n} = \max_{G_i\in \mathcal{G}}(|\mathcal{V}^{G_i}|)$) due to the total $|\mathcal{G}|$ graphs. After computing the pairwise similarity, we can construct the GoG by naively thresholding out the top$-k$ similar graphs for each graph and the time complexity here is $O(|\mathcal{G}|k)$. By default $k\le |\mathcal{G}|$, we directly have $O(|\mathcal{G}|k) < O(|\mathcal{G}|^2) = O({|\mathcal{G}| \choose 2}) < O({|\mathcal{G}| \choose 2}\tilde{n}^3)$ and hence the time complexity of the first module is $O({|\mathcal{G}| \choose 2}\tilde{n}^3)$. Despite the prohibitively heavy computation of $O({|\mathcal{G}| \choose 2}\tilde{n}^3)$, the whole module is a pre-procession computation once for all and we can further save the already computed similarity matrix $\mathbf{S}$ for future use, which therefore imposes no computational challenge. In topological augmentation, we augment graphs $T$ times during each training epoch and each time we either go over all its edges or nodes, therefore the total time complexity of this module during each training epoch is $O(T\sum_{G_i\in\mathcal{G}^B}(|\mathcal{V}^{G_i}|+|\mathcal{E}^{G_i}|))$. Since augmenting graphs multiple times gains no further improvement than 2~\cite{graphCL}, we fix $T$ to be the constant 2 and therefore the total complexity of this part is linearly proportional to the size of each graph, which imposes no additional time compared with GNN encoders. Among the GoG propagation component, the most  computational part comes from propagation in Eq.~\eqref{eq:smoothing}, which can be efficiently computed by applying power iteration from the edge view in $O(K|\mathcal{E}^{\mathcal{G}^{\text{kNN}, B}}|)$ for each subgraph induced by graphs in batch $\mathcal{G}^B$. Based on experimental results in Figure~\ref{fig-MTUAGKNNK}-\ref{fig-DHFRKNNK}, we usually choose $k$ to be small to ensure the sparcity and the high homophily of GoG, then $O(K|\mathcal{E}^{\mathcal{G}^{\text{kNN}, B}}|)$ can be neglected compared with applying GNN encoders to get representations of each graph, $O(K{\sum_{G_i\in\mathcal{G}^{B}}{|\mathcal{E}^{G_i}|}})$. For the model complexity, besides the parameters of GNN encoders, G$^2$GNN adds %introduces 
no additional parameters and therefore its model complexity is exactly the same as traditional GNN encoders. % such as GIN.

% Among the GoG propagation component, the most  computational part comes from propagation in Eq.~\eqref{eq:smoothing}, which can be efficiently computed by applying power iteration from the edge view in $O(K|\mathcal{E}^{\mathcal{G}^{\text{kNN}, B}}|)$ for each subgraph induced by graphs in batch $\mathcal{G}^B$. Based on experimental results in Figure~\ref{fig-MTUAGKNNK}-\ref{fig-DHFRKNNK}, we usually choose $k$ to be small to ensure the sparcity and the high homophily of GoG, then $O(K|\mathcal{E}^{\mathcal{G}^{\text{kNN}, B}}|)$ can be neglected compared with applying GNN encoders to get representations of each graph, $O(K{\sum_{G_i\in\mathcal{G}^{B}}{|\mathcal{E}^{G_i}|}})$.

% For the model complexity, besides the parameters of GNN encoders, G$^2$GNN adds %introduces 
% no additional parameters and therefore its model complexity is exactly the same as traditional GNN encoders. % such as GIN.

In summary, our model introduces no extra model complexity but $O({|\mathcal{G}| \choose 2}\tilde{n}^3)$ extra time complexity in the pre-procession stage. We further presents the actual time used for applying Shortest Path kernel to compute $\mathbf{S}$ in Table~\ref{table-statist-dataset}. It can be clearly see that similarity matrix $\mathbf{S}$ is calculated in a short time for each dataset other than D\&D and REDDIT-B since graphs in these two dataset are on average denser than other datasets as shown in Table~\ref{table-statist-dataset}. However, we can simply pre-compute this $\mathbf{S}$ once for all and reuse it for G$^2$GNN. Moreover, we can make this computation feasible by either employing the fast shortest-path kernel computations by sampling-based approximation where we sample pairs of nodes and compute shortest paths between them~\cite{kilhamn2015fast}.
% constructing the graph of graphs via other representation learning techniques such as graph neural networks.

%\vspace{-1ex}
%\vspace{1.5ex}
\section{Evaluation}\label{sec: experiment}
In this section, we evaluate the effectiveness of G$^2$GNN by conducting extensive imbalanced graph classification on multiple types of graph datasets with different levels of imbalance. We begin by introducing the experimental setup, including datasets, baselines, evaluation metrics, and parameter settings.

\begin{table}[t]%[htbp!]
\footnotesize
\centering
% \vskip -5.25ex
\setlength{\extrarowheight}{.095pt}
\setlength\tabcolsep{4pt}
\caption{Statistics of datasets}
\vspace{-2.5ex}
\label{table-statist-dataset}
\begin{tabular}{lccccc}
\Xhline{2\arrayrulewidth}%\hline
\textbf{Networks} & \# \textbf{Graphs} & \# \textbf{Avg-Node} & \# \textbf{Avg-Edge} & \# \textbf{Attr} & \textbf{Time}(s)*   \\
\Xhline{1.5\arrayrulewidth}
PTC-MR~\cite{toivonen2003statistical} & $344$ & $14.29$ & $14.69$ & $18$ & $0.257$\\
NCI1~\cite{wale2008comparison} & $4110$	& $29.87$	& $32.30$ & $37$ & $11.21$\\
MUTAG~\cite{debnath1991structure} & $188$ & $17.93$ & $19.79$ & $7$ & $0.212$\\
PROTEINS~\cite{yanardag2015deep} & $1113$ & $39.06$ & $72.82$ & $3$ & $11.36$\\
D\&D~\cite{sutherland2003spline} & $1178$	& $284.32$ & $715.66$ & $89$ & $574.71$\\
DHFR~\cite{sutherland2003spline} & $756$ & $42.43$ & $44.54$ & $3$ & $3.70$\\
REDDIT\-B~\cite{yanardag2015deep} & $2000$ & $429.63$ & $497.75$ & $\backslash$ & $3376$\\
\Xhline{2\arrayrulewidth}
\end{tabular}
\vskip 1ex
\begin{tablenotes}
      \footnotesize
      \item \textbf{*}  The column 'time' represents the actual time used for applying Shortest Path kernel to compute $\mathbf{S}$ for each dataset.
\end{tablenotes}
\vskip 1ex
%\vspace{1ex}
\end{table}

\vspace{-1ex}
\subsection{Experimental Setup}\label{sec-experimentsetup}
\subsubsection{Dataset} 
We conduct experiments on seven widely-adopted real-world datasets~\cite{yanardag2015deep,sutherland2003spline}, which include:
\begin{inparaenum}
    \item Chemical compounds: 
    PTC-MR,
    NCI1,
    and MUTAG.
    \item Protein compounds: 
    PROTEINS,
    D\&D,
    and DHFR.
    \item Social network: 
    REDDIT-B.
\end{inparaenum} 
Details of these datasets can be found in Table~\ref{table-statist-dataset}.

\vspace{-1ex}
\subsubsection{Baselines}
To evaluate the effectiveness of the proposed G$^2$GNN, we select three models designed for graph classification, which includes:
% \begin{inparaenum}
%     \item \textbf{GIN}~\cite{powerfulgnn}: A basic supervised GNN model for graph classification due to its distinguished expressiveness of graph topology. 
%     \item \textbf{InfoGraph}~\cite{Infograph}: An unsupervised GNN model for learning graph representations via maximizing the mutual information between the whole graph and its substructures of different scales.
%     \item \textbf{GraphCL}~\cite{graphCL}: Stepping further from InfoGraph, GraphCL proposes four strategies to augment graphs and learns graph representations by maximizing the mutual information between the original graph and its augmented variants.
% \end{inparaenum}

\begin{itemize}[leftmargin=*]
    \item \textbf{GIN}~\cite{powerfulgnn}: A basic supervised GNN model for graph classification due to its distinguished expressiveness of graph topology. 
    \item \textbf{InfoGraph}~\cite{Infograph}: An unsupervised GNN model for learning graph representations via maximizing mutual information between the original graph and its substructures of different scales.
    \item \textbf{GraphCL}~\cite{graphCL}: Stepping further from InfoGraph, GraphCL proposes four strategies to augment graphs and learns graph representations by maximizing the mutual information between the original graph and its augmented variants.
\end{itemize}

Since imbalanced datasets naturally provide weak supervision on minority classes, unsupervised GNNs outweigh supervised counterparts and selecting them as baselines could more confidently justify the superiority of our model. All the above three baselines are proposed without consideration of imbalanced setting, therefore we further equip these three backbones with strategies designed 

\noindent specifically for handling imbalance issue, which includes:
\begin{itemize}[leftmargin=*]
    \item \textbf{Upsampling} (\textit{us}): A classical approach that repeats samples from minority classes~\cite{kubat1997addressing}. We implement this directly in the input space by duplicating minority graphs.
    \item \textbf{Reweighting} (\textit{rw}): %
    A general cost-sensitive approach introduced in~\cite{yuan2012sampling+} that assigns class-specific weights in computing the classification loss term in Eq.~\eqref{eq: obj}; we set the weights of each class as inverse ratio of the total training graphs to the number of training graphs in that class.
    \item \textbf{SMOTE} (\textit{st}): Based on the ideas of SMOTE~\cite{smote}, synthetic minority samples are created by interpolating minority samples with their nearest neighbors within the same class based on the output of last GNN layer. Since directly interpolating in the topological space may generate invalid graph topology, here we first obtain graph representations by GNN-based encoders and interpolate minority graph representations in the embedding space to generate more minority training instances. Here, the nearest neighbors are computed according to Euclidean distance.
\end{itemize}

% \begin{inparaenum}
%     \item \textbf{Upsampling (us)}: A classical approach that repeats samples from minority classes~\cite{kubat1997addressing}. We implement this directly in the input space by duplicating minority graphs.
%     \item \textbf{Reweighting (rw)}: %
%     A general cost-sensitive approach introduced in~\cite{yuan2012sampling+} that assigns class-specific loss weights in computing the classification loss term in Eq.~\eqref{eq: obj}; we set the weights of each class as inverse ratio of the total training graphs to the number of training graphs in that class.
%     \item \textbf{SMOTE (st)}: Based on the ideas of SMOTE~\cite{smote}, synthetic minority samples are created by interpolating minority samples with their nearest neighbors within the same class based on the output of last GNN layer. Since directly interpolating in the topological space may generate invalid graph topology, here we first obtain graph representations by GNN-based encoders and interpolate minority graph representations in the embedding space to generate more minority training instances. Specifically, nearest neighborhoods are computed according to Euclidean distance.
% \end{inparaenum}

\noindent Equipping each of the above three backbones with up-sampling, re-weighting, and SMOTE strategies tailored specifically for imbalanced classification, we end up with 10 baselines. Specifically, we equip up-sampling and re-weighting with all three backbones and name each new baseline by combining the name of its backbone and the equipped strategy, e.g, $\text{GIN}_{us}$ represents the backbone $\text{GIN}$ equipped with the up-sampling strategy. Since applying SMOTE empirically leads to similar or even worse performance gains, we only stack it on the GIN backbone.

\begin{table*}[t]
\footnotesize
\setlength\tabcolsep{5pt}
\renewcommand{\arraystretch}{1}
\caption{Graph classification performance on seven datasets. Note that the standard deviation is relatively higher since we focus on the imbalance problem and use 50 different data splits (i.e., having different training data distributions). %We equip each backbone GIN, InfoGraph and GraphCL with $us, rw, st$ corresponding to the up-sampling, re-weight, and SMOTE strategy, respectively. 
G$^2$GNN$_e$ and G$^2$GNN$_n$ represent our proposed model using the removing edges and masking node features augmentation strategy, respectively. 
Red (blue) denotes the \textcolor{red}{best} (\textcolor{blue}{runner-up}) model.}
%Here red represents the best one and blue denotes the runner-up}
    \vskip -1.5ex
\label{table-result}
%\raggedright \hspace{13ex}
\begin{tabular}{lcccccccc}
\hline
\multirow{2}{*}{\textbf{Model}} & \multicolumn{2}{c}{\textbf{MUTAG} (5:45)} & \multicolumn{2}{c}{\textbf{PROTEINS} (30:270)} & \multicolumn{2}{c}{\textbf{D\&D} (30:270)} & \multicolumn{2}{c}{\textbf{NCI1} (100:900)} \\
 & F1-macro & F1-micro & F1-macro & F1-micro & F1-macro & F1-micro & F1-macro & F1-micro \\
\hline
GIN & $52.50\pm 18.70$ & $56.77\pm 14.14$ & $25.33\pm 7.53$ & $ 28.50\pm 5.82$ & $9.99\pm 7.44$ & $11.88\pm 9.49$ & $18.24\pm 7.58$ & $18.94 \pm 7.12$ \\
GIN$_{us}$ & $78.03\pm 7.62$ & $78.77\pm 7.67$ & $65.64\pm 2.67$ & $71.55\pm 3.19$ & $41.15\pm 3.74$ & $70.56\pm 10.28$ & $59.19\pm 4.39$ & $71.80\pm 7.02$ \\
GIN$_{rw}$ & $77.00\pm 9.59$ & $77.68\pm 9.30$ & $54.54\pm 6.29$ & $55.77\pm 7.11$ & $28.49\pm 5.92$ & $40.79\pm 11.84$ & $36.84\pm 8.46$ & $39.19\pm 10.05$ \\
GIN$_{st}$ & $74.61\pm 9.66$ & $75.11\pm 9.87$ & $56.07\pm 7.95$ & $57.85\pm 8.70$ & $27.08\pm 8.63$ & $39.01\pm 15.87$ & $40.40\pm 9.63$ & $44.48\pm 12.05$ \\
 \hline
InfoGraph & $69.11\pm 9.03$ & $69.68\pm 7.77$ & $35.91\pm 7.58$ & $36.81\pm 6.51$ & $21.41\pm 4.51$ & $27.68\pm 7.52$ & $33.09\pm 3.30$ & $34.03\pm 3.68$ \\
InfoGraph$_{us}$ & $78.62\pm 6.84$ & $79.09\pm 6.86$ & $62.68\pm 2.70$ & $66.02\pm 3.18$ & $41.55\pm 2.32$ & $71.34\pm 6.76$ & $53.38\pm 1.88$ & $62.20\pm 2.63$ \\
InfoGraph$_{rw}$ & \textcolor{blue}{$80.85\pm 7.75$} & \textcolor{blue}{$81.68\pm 7.83$} & $65.73\pm 3.10$& $69.60\pm 3.68$ & $41.92\pm 2.28$ & $72.43\pm 6.63$ & $53.05\pm 1.12$ & $62.45\pm 1.89$\\
 \hline
GraphCL & $66.82\pm 11.56$ & $67.77\pm 9.78$ & $40.86\pm 6.94$  &  $41.24\pm 6.38$& $21.02\pm 3.05$ & $26.80\pm 4.95$ & $31.02\pm 2.69$ & $31.62\pm 3.05$ \\
GraphCL$_{us}$ & $80.06\pm 7.79$ & $80.45\pm 7.86$ &$64.21\pm 2.53$  & $65.76\pm 2.61$ & $38.96\pm 3.01$ & $64.23\pm 8.10$ & $49.92\pm 2.15$ & $58.29\pm 3.30$ \\
GraphCL$_{rw}$ & $80.20\pm 7.27$ & $80.84\pm 7.43$ & $63.46\pm 2.42$ & $64.97\pm 2.41$ & $40.29\pm 3.31$ & $67.96\pm 8.98$ & $50.05\pm 2.09$ & $58.18\pm 3.08$ \\
 \hline
G$^2$GNN$_{e}$ & $80.37\pm 6.73$ & $81.25\pm 6.87$ & \textcolor{red}{$67.70\pm 2.96$} & \textcolor{blue}{$73.10\pm 4.05$} & \textcolor{blue}{$43.25\pm 3.91$} & \textcolor{blue}{$77.03\pm 9.98$} & \textcolor{blue}{$63.60\pm 1.57$} & \textcolor{blue}{$72.97\pm1.81$}\\
G$^2$GNN$_{n}$ & \textcolor{red}{$83.01\pm 7.01$} & \textcolor{red}{$83.59\pm 7.14$} & \textcolor{blue}{$67.39\pm 2.99$} & \textcolor{red}{$73.30\pm 4.19$} & \textcolor{red}{$43.93\pm 3.46$} & \textcolor{red}{$79.03\pm 10.78$} & \textcolor{red}{$64.78\pm 2.86$} & \textcolor{red}{$74.91\pm 2.14$}\\
 \hline
\end{tabular}
\vskip 0.5ex
%\hspace{13ex}
\begin{tabular}{llcccccccc}
\hline
\multirow{2}{*}{\textbf{Model}} & \multicolumn{2}{c}{\textbf{PTC-MR} (9:81)} & \multicolumn{2}{c}{\textbf{DHFR} (12:108)} & \multicolumn{2}{c}{\textbf{REDDIT-B} (50:450)} & \multicolumn{2}{c}{\textbf{Ave. Rank}} \\
& F1-macro & F1-micro & F1-macro & F1-micro & F1-macro & F1-micro & F1-macro & F1-micro \\
\hline
GIN & $17.74\pm 6.49$ & $20.30\pm 6.06$ & $35.96\pm 8.87$ & $49.46\pm 4.90$ & $33.19\pm 14.26$ & $36.02\pm 17.38$ & 12.00 & 12.00 \\
GIN$_{us}$ & $44.78\pm 8.01$ & $55.43\pm 14.25$ & $55.96\pm 10.06$ & $59.39\pm 6.52$ & $66.71\pm 3.92$ & $83.00\pm 5.18$ & 5.00 & 4.43 \\
GIN$_{rw}$ & $36.96\pm 14.08$ & $43.09\pm 20.01$ & $55.16\pm 9.47$ & $57.78\pm 6.69$ & $45.17\pm 8.46$ & $51.92\pm 12.29$ & 8.86 & 8.86 \\
GIN$_{st}$ & $36.30\pm 11.45$ & $40.04\pm 15.32$ & $56.06\pm 9.60$ & $58.48\pm 6.42$ & $60.05\pm 4.14$ & $73.59\pm 6.05$ & 8.29 & 8.43 \\
\hline
InfoGraph & $25.85\pm 6.14$ & $26.71\pm 6.50$ & $50.62\pm 8.33$ & $56.28\pm 4.58$ & $57.67\pm 3.80$ & $67.10\pm 4.91$ & 10.00 & 10.14 \\
InfoGraph$_{us}$ & $44.29\pm 4.69$ & $48.91\pm 7.49$ & $59.49\pm 5.20$ & $61.62\pm 4.18$ & $67.01\pm 3.34$ & $78.68\pm 3.71$ & 5.00 & 5.00\\
InfoGraph$_{rw}$ & $44.09\pm 5.62$ & $49.17\pm 8.78$  & $58.67\pm 5.82$ & $60.24\pm 4.80$ & $65.79\pm 3.38$ & $77.35\pm 3.96$ & \textcolor{blue}{4.43} & 4.29\\
\hline
GraphCL & $24.22\pm 6.21$ & $25.16\pm 5.25$ & $50.55\pm 10.01$ & $56.31\pm 6.12$ & $53.40\pm 4.06$ & $62.19\pm 5.68$ & 10.71 & 10.57 \\
GraphCL$_{us}$ & $45.12\pm 7.33$ & $53.50\pm 13.31$ & $60.29\pm 9.04$ & $61.71\pm 6.75$ & $62.01\pm 3.97$ & $75.84\pm 3.98$ & 5.29 & 5.43 \\
GraphCL$_{rw}$ & $44.75\pm 7.62$ & $52.22\pm 13.24$ & \textcolor{blue}{$60.87\pm 6.33$} & \textcolor{blue}{$61.93\pm 5.15$} & $62.79\pm 6.93$ & $76.15\pm 9.15$ & 5.00 & 5.29\\
\hline
G$^2$GNN$_e$ & \textcolor{blue}{$46.40\pm 7.73$}  & \textcolor{blue}{$56.61\pm13.72$} & \textcolor{red}{$61.63\pm 10.02$} & \textcolor{red}{$63.61\pm 6.05$} & \textcolor{red}{$68.39\pm 2.97$} & \textcolor{red}{$86.35\pm 2.27$} & \textcolor{red}{1.71}  & \textcolor{blue}{1.86}\\
G$^2$GNN$_n$ & \textcolor{red}{$46.61\pm 8.27$} & \textcolor{red}{$56.70\pm 14.81$} & $59.72\pm 6.83$ & $61.27\pm 5.40$ & \textcolor{blue}{$67.52\pm 2.60$} & \textcolor{blue}{$85.43\pm 1.80$} & \textcolor{red}{1.71} & \textcolor{red}{1.71}\\
 \hline
\end{tabular}
    \vskip -2ex
\end{table*}

\vspace{-1.5ex}
\subsubsection{Evaluation Metrics}
Following existing work in imbalanced classification~\cite{graphsmote}, we use two criterion: F1-macro and F1-micro to measure the performance of G$^2$GNN and other baselines. F1-macro computes the accuracy independently for each class and then takes the average (i.e., treating different classes equally). F1-micro computes accuracy over all testing examples at once, which may underweight the minority classes. Following~\cite{GRAND}, The whole GoG propagation is conducted in the transductive setting where representations of graphs in the training set could aggregate representations of graphs in the validation and testing sets while the classification loss is only evaluated on the given training labels.

\vspace{-1.5ex}
\subsubsection{Parameter Settings}
We implement our proposed G$^2$GNN and some necessary baselines using Pytorch Geometric~\cite{fey2019fast}. For InfoGraph\footnote{\url{https://github.com/fanyun-sun/InfoGraph}} and  GraphCL\footnote{\url{https://github.com/Shen-Lab/GraphCL}} we use the original authors' code with any necessary modifications. Aiming to provide a rigorous and fair comparison across models on each dataset, we tune hyperparameters for all models individually as: the weight decay $\in[0, 0.1]$, the encoder hidden units $\in\{128, 256\}$, the learning rate $\in\{0.001, 0.01\}$, the inter-network level propagation $L\in \{1, 2, 3\}$, the augmentation ratio $\delta \in \{0.05, 0.1, 0.2\}$, the number of neighboring graphs in constructing GoG $k\in\{2, 3, 4\}$, the augmentation number $T = 2$ and sharpening temperature $\tau = 0.5$. We employ Shortest Path Kernel to compute similarity matrix $\mathbf{S}$ and set the trainable classifier $g$ as a 2-layer MLP. For REDDITB dataset, we use one-hot encoding of the node degree as the feature of each node following~\cite{graphCL, Infograph}. For reproducibility, the code of the model with its corresponding hyperparameter configurations are publicly available\footnote{Code for G$^2$GNN: \href{https://github.com/submissionconff/G2GNN}{\textcolor{blue}{https://github.com/submissionconff/G2GNN}}}.

\begin{figure*}[t]
     \centering
     %\vskip -1ex
     \includegraphics[width=.85\textwidth]{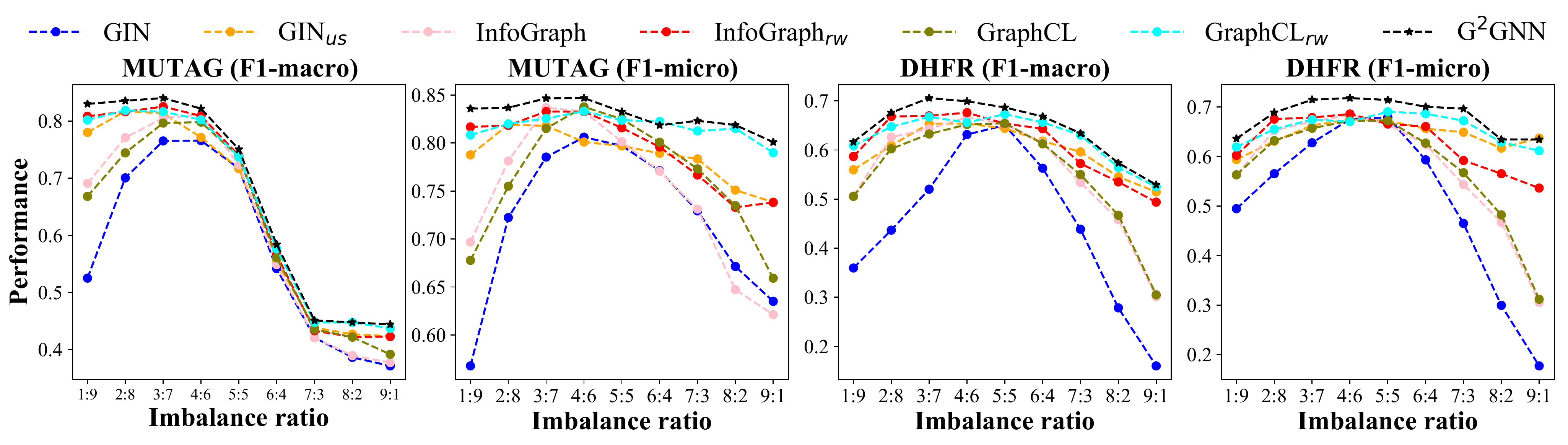}
     \vskip -2ex
     \caption{Graph classification results under different class imbalance ratios where 5:5 corresponds to balanced scenario while 1:9 and 9:1 correspond to highly imbalance scenario. Compared with GIN(\textcolor{blue}{blue}), InfoGraph(\textcolor{pink}{pink}), GraphCL(\textcolor{olive}{olive}) designed not specifically for imbalanced scenario, our G$^2$GNN(black) model outperforms all of them in nearly all imbalance ratio settings and the margin further increases as the level of imbalance increases (i.e., deviates from the balanced scenario). Note that here we use the same amount of training and validation graphs (25\%/25\%) as used in Table~\ref{table-result}.}

     \label{fig-imb_ratio}
     \vskip -2.5ex
\end{figure*}

\vspace{-1ex}
\subsection{Performance Comparison}\label{subsec-pc}
In this subsection, we compare the performance of \textbf{G$^2$GNN}$_{e}$ and \textbf{G$^2$GNN}$_{n}$, which represent the G$^2$GNN framework with the edge removal or node feature masking as augmentation, respectively, against the aforementioned baselines. Since class distributions of most datasets are not strictly imbalanced, we use an imitative imbalanced setting: we randomly set 25\%/25\% graphs as training/validation sets and among each of them, we choose one class as minority and reduce the graphs of this class in the training set (increase the other one) till the imbalance ratio reaches 1:9, which creates an extremely imbalanced scenario\footnote{We select the amount of training and validation data as 25\% to ensure the sufficiency of minority instances in both training and validation set given the imitative data distribution is at such a skewed level}. We average the performance per metric across 50 different data splits to avoid any bias from data splitting. Table~\ref{table-result} reports the mean and the standard deviation of the performance.

We observe from Table~\ref{table-result} that G$^2$GNN performs the best in all 7 datasets under both F1-macro and F1-micro. Moreover, edge removing (i.e., G$^2$GNN$_e$) benefits more on the social network (i.e., REDDIT-B) while node feature masking (i.e., G$^2$GNN$_n$) enhances more on biochemical molecules (e.g., MUTAG, D\&D, NCI1 and PTC-MR), which conforms to~\cite{graphCL} and is partially attributed to no node attributes presented in the social network. Models that are specifically designed for tackling the class imbalance issue generally perform better than the corresponding bare backbones without any strategy handling imbalance. The inferior performance of GIN$_{rw(st)}$ to GIN$_{us}$ is because we either set weights for adjusting training loss of different classes or generate synthetic samples based on training data at current batch. Since the number of training instances in each batch may not strictly follow the prescribed imbalance ratio, the batch-dependent weight or synthetic samples hardly guarantee the global balance. InfoGraph(GraphCL)-based variants do not suffer from the issue introduced by batch-training since once we obtain graph representations from pre-trained models by mutual information maximization, we feed them through downstream classifiers all at once without any involvement of batch process. Therefore, the performance of InfoGraph(GraphCL)$_{rw(st)}$ is comparable to InfoGraph(GraphCL)$_{us}$. We emphasize that the larger standard deviation in our setting is due to the significantly different training data across different runs. We further argue that this standard deviation cannot be reduced by only increasing the number of runs due to the imbalance nature of the problem. However, the higher average performance of our model still signifies its superiority in handling a wide range of imbalanced data splittings.

\vspace{-1.25ex}
\subsection{Influence of Imbalance Ratio} 
We further compare the performance of our model with other baselines under different imbalance ratios. We vary the imbalance ratio from 1:9 to 9:1 by fixing the total number of training and validation graphs as 25\%/25\% of the whole dataset as before and gradually varying the number of graphs in different classes, which exhausts the imbalance scenarios from being balanced (5:5) to the extremely imbalanced (1:9 or 9:1) scenarios. Note that for clear comparison, we only visualize the performance of the best variant among each of three backbones in Figure~\ref{fig-imb_ratio}. We can clearly see that the performance of all models first increases and then decreases as the imbalance ratio increases from 0.1 to 0.9, which demonstrates the detrimental effect of data imbalance on the model performance and such detrimental effect becomes even worse when the imbalance becomes more severe. Furthermore, the F1-macro score of our G$^{2}$GNN model clearly outperforms all other baselines on both MUTAG and DHFR under each imbalance ratio, which soundly justifies the superiority and robustness of our model in alleviating imbalance of different level. Different from supervision presented from given labeled data, the extra supervision derived by leveraging neighboring graphs' information via propagation and topological augmentation is weakly influenced by the amount of training data. Therefore, the margin achieved by our model further grows when imbalance ratio is either too low or too high compared with GIN, InfoGraph and GraphCL that are not designed specifically for handling the imbalance scenario since the extra supervision derived in our model stays the same while the basic supervision encoded in the training data decreases. Besides, our model also performs comparable or even slightly better then all other baselines under balanced scenario, which additionally signifies the potentiality of our model in balanced data-splitting. Among other baselines, GraphCL$_{rw}$ performs the best since it applies re-weight strategy to balance the training loss and further leverages the graph augmentation coupled with mutual information maximization to extract the most relevant information for downstream classification. An interesting observation is that the optimal performance is not always %achieved
when the labeled data is strictly balanced, which reflects the uneven distribution of informatic supervision embedded across different classes. 

\vspace{-2ex}
\subsection{Ablation Study}
\begin{figure}[t]
     \centering
     %\vskip -2ex
     \includegraphics[width=0.48\textwidth]{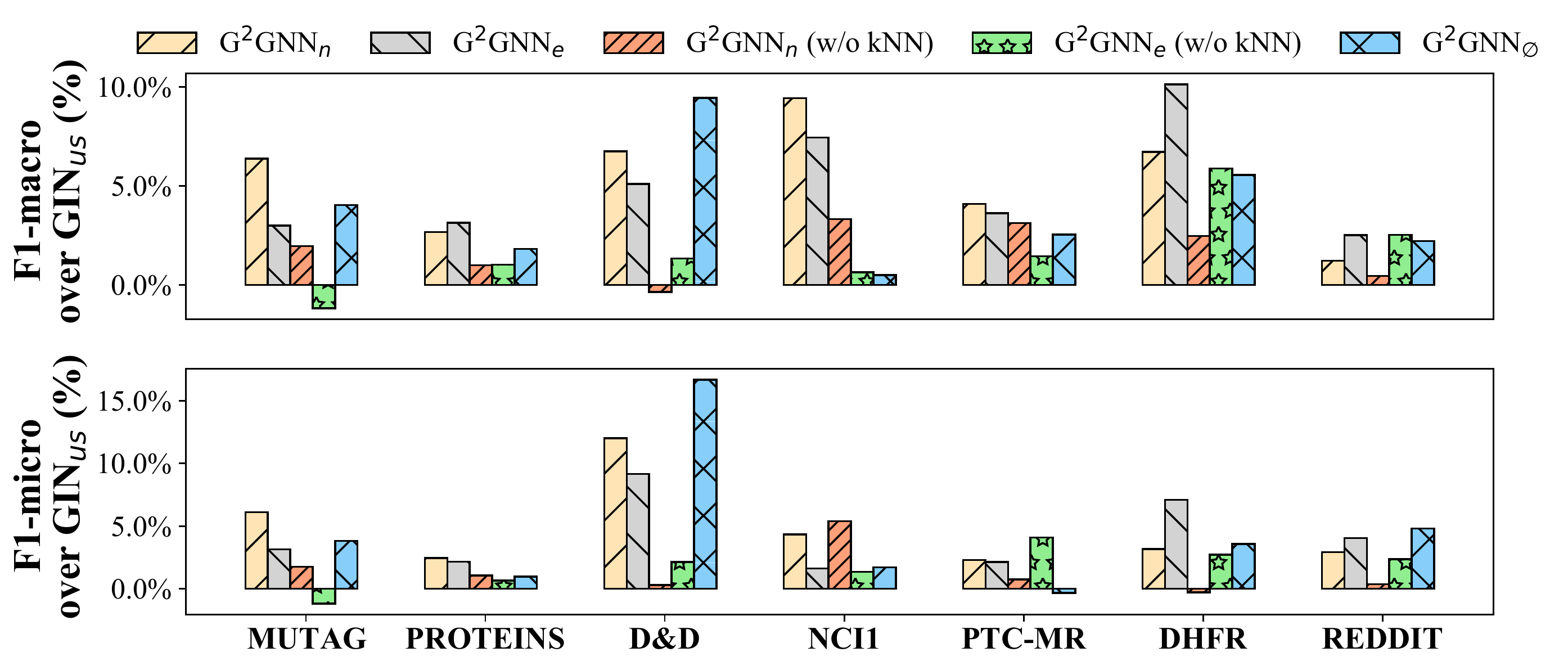}
     \vskip -2ex
     \caption{Ablation study results of G$^2$GNN where we report the improvement over GIN$_{us}$ due to its simplicity and effectiveness %as a baseline 
     (seen in Table~\ref{table-result}) for understanding relative improvements of each G$^2$GNN component.}
     \label{fig-ablation}
     \vskip 1ex
     \vspace{-1ex}
\end{figure}

In this section, we conduct ablation study to fully understand the effect of each component in G$^2$GNN on alleviating the imbalance issue. In Figure~\ref{fig-ablation}, we present performance improvement over the baseline GIN$_{us}$ achieved by our proposed framework (\textbf{G$^2$GNN$_{e(n)}$}) along with variants that remove the GoG propagation (\textbf{G$^2$GNN$_{e(n)}$ (w/o kNN)}) and remove the topological augmentation (\textbf{G$^2$GNN$_{\emptyset}$}). 
\begin{inparaenum}
\textbf{(1)} We notice that solely employing GoG propagation (\textbf{G$^2$GNN$_{\emptyset}$}) increases the performance on all datasets according to F1-macro, demonstrating the effectiveness of GoG propagation in alleviating imbalance issue.
\textbf{(2)} Augmenting via removing edges hurts the performance on MUTAG. This is because the size of each graph in MUTAG is relatively small and thus removing edges may undermine crucial topological information related to downstream classification.
\textbf{(3)} We observe that the proposed GoG propagation and graph augmentation generally achieve more performance boost on F1-macro than F1-micro. This is because the derived supervision significantly enhance the generalizability of training data in minority classes. However, for majority classes where majority training instances already guarantee high generalizability, the enhancement would be minor.
\textbf{(4)} Combining GoG propagation and graph augmentation together is better than only applying one of them in most cases, which indicates that the extra supervision derived by globally borrowing neighboring information and locally augmenting graphs are both beneficial to downstream tasks and not overlapped with each other as the accumulating benefit shown here.
\textbf{(5)} On NCI1, despite the minor improvement of applying only one of the proposed two modules, combining them together leads to significant progress. This is because instead of propagating original graphs' representations, we leverage augmented graphs in GoG propagation and the derived local supervision is further enhanced by the global propagation to create more novel supervision and extremely enhance the model generalibility on minority classes.
\end{inparaenum}

\vspace{-1.5ex}
\subsection{Further Probe}
%\subsection{Sensitive analysis}
\subsubsection{Effect of Neighborhood Numbers}
\begin{figure}[t]%[htbp!]
     %\vspace{-2ex}
     \centering
     \begin{subfigure}[b]{0.235\textwidth}
         \centering
         \includegraphics[width=0.98\textwidth]{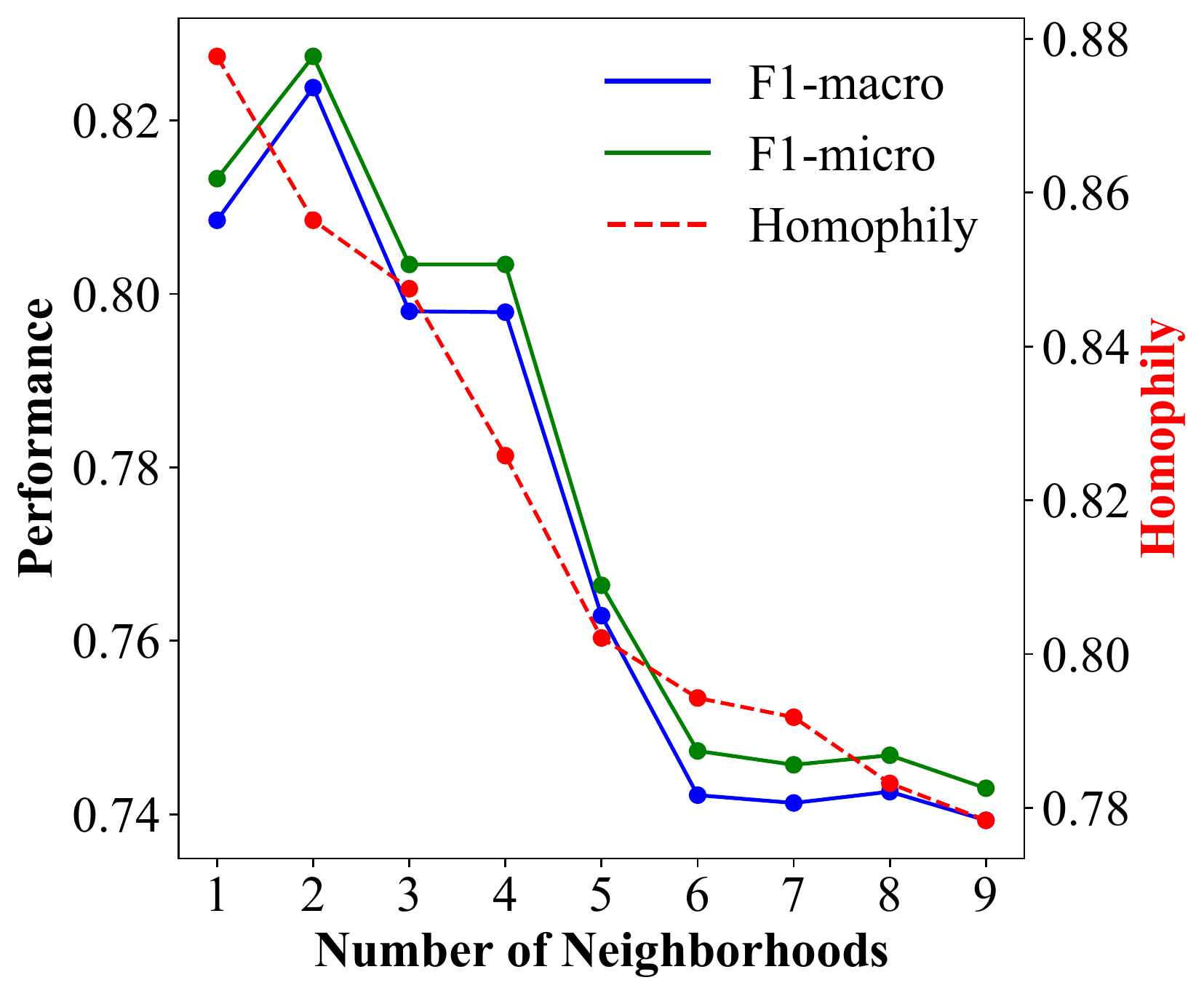}
         \caption{MUTAG}
         \label{fig-MTUAGKNNK}
     \end{subfigure}
     \begin{subfigure}[b]{0.235\textwidth}
         \centering
         \includegraphics[width=0.98\textwidth]{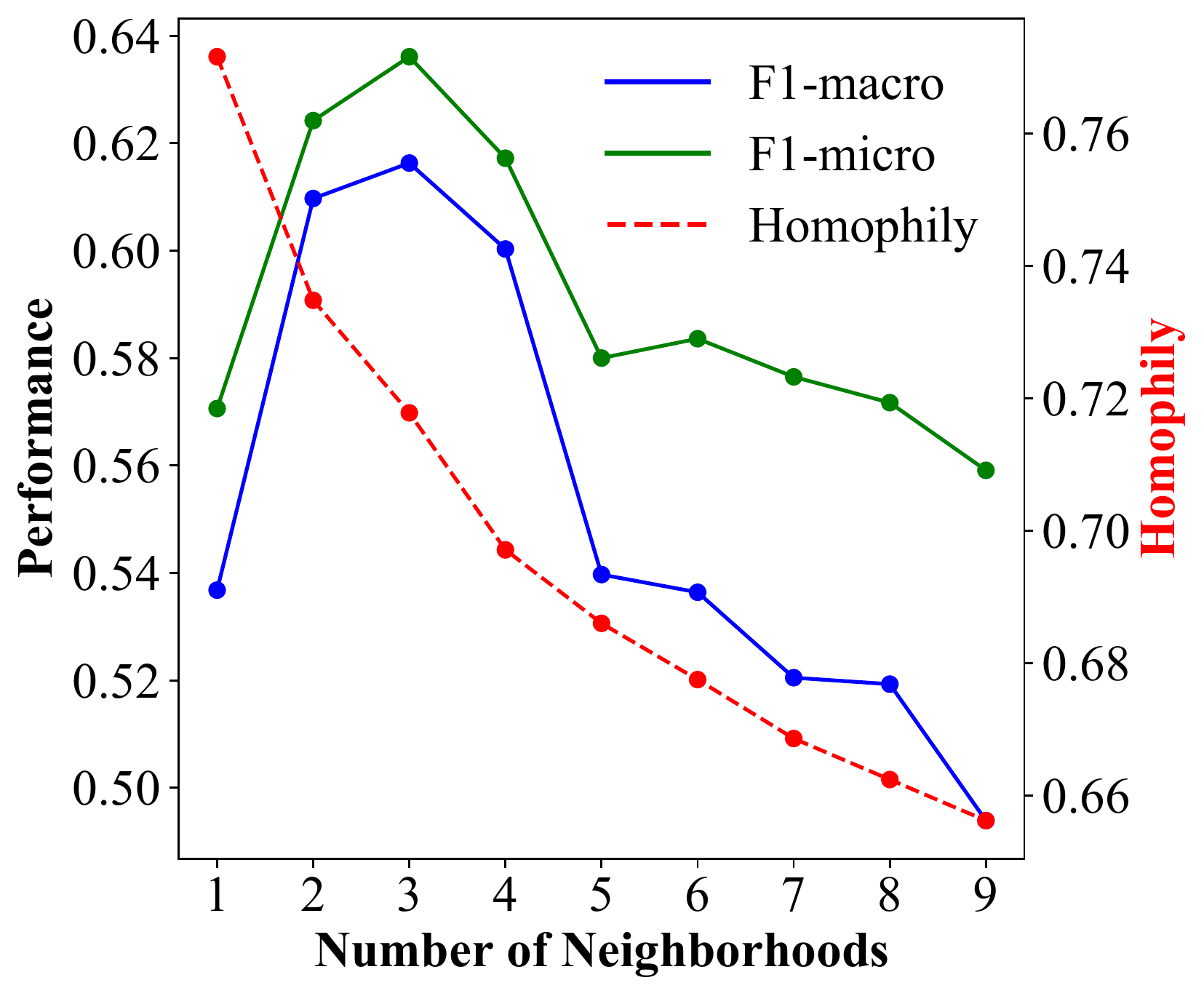}
         \caption{DHFR}
         \label{fig-DHFRKNNK}
     \end{subfigure}
     \vskip -2ex
    %  \vspace{-1.5ex}
     \caption{Relationship between neighborhood number, edge homophily, and performance on MUTAG and DHFR. The performance first increases and then decreases as the number of neighborhoods increases on $\mathcal{G}^{\text{kNN}}$. The reported result here is averaged over 20 runs.}
     \label{fig-layer}
\end{figure}

Here we investigate the influence of the number of neighboring graphs on the performance of G$^2$GNN$_n$ on MUTAG and DHFR. The experimental setting is the same as Section~\ref{sec-experimentsetup} except that we alter the $k$ among $\{1, 2, ..., 9\}$. In Figure~\ref{fig-layer}, we see that both of the F1-macro and F1-micro increase first as $k$ increases to 2 on MUTAG and 3 on DHFR since higher $k$ means more number of neighboring graphs sharing the same label, as the homophily level at this stage is generally higher given the red line, therefore we derive more beneficial supervision. However, as we further increases $k$ to 6, the performance begins to decrease since most of added neighborhoods share different labels due to low homophily in this middle stage and hence provide adverse information that compromises classification. In the last stage when $k$ proceeds to increase beyond $6$, the performance gradually becomes stable, this is because directly linking each graph with its 6-top similar graphs leads to a very dense GoG and propagation on this dense GoG directly incorporates information from most of other graphs and therefore the neighboring information that each graph receives is too noisy and useless.

\begin{figure}[t]%[htbp!]
     %\vspace{-2ex}
     \vspace{0.5ex}
     \centering
     \begin{subfigure}[b]{0.235\textwidth}
         \centering
         \includegraphics[width=1.00\textwidth]{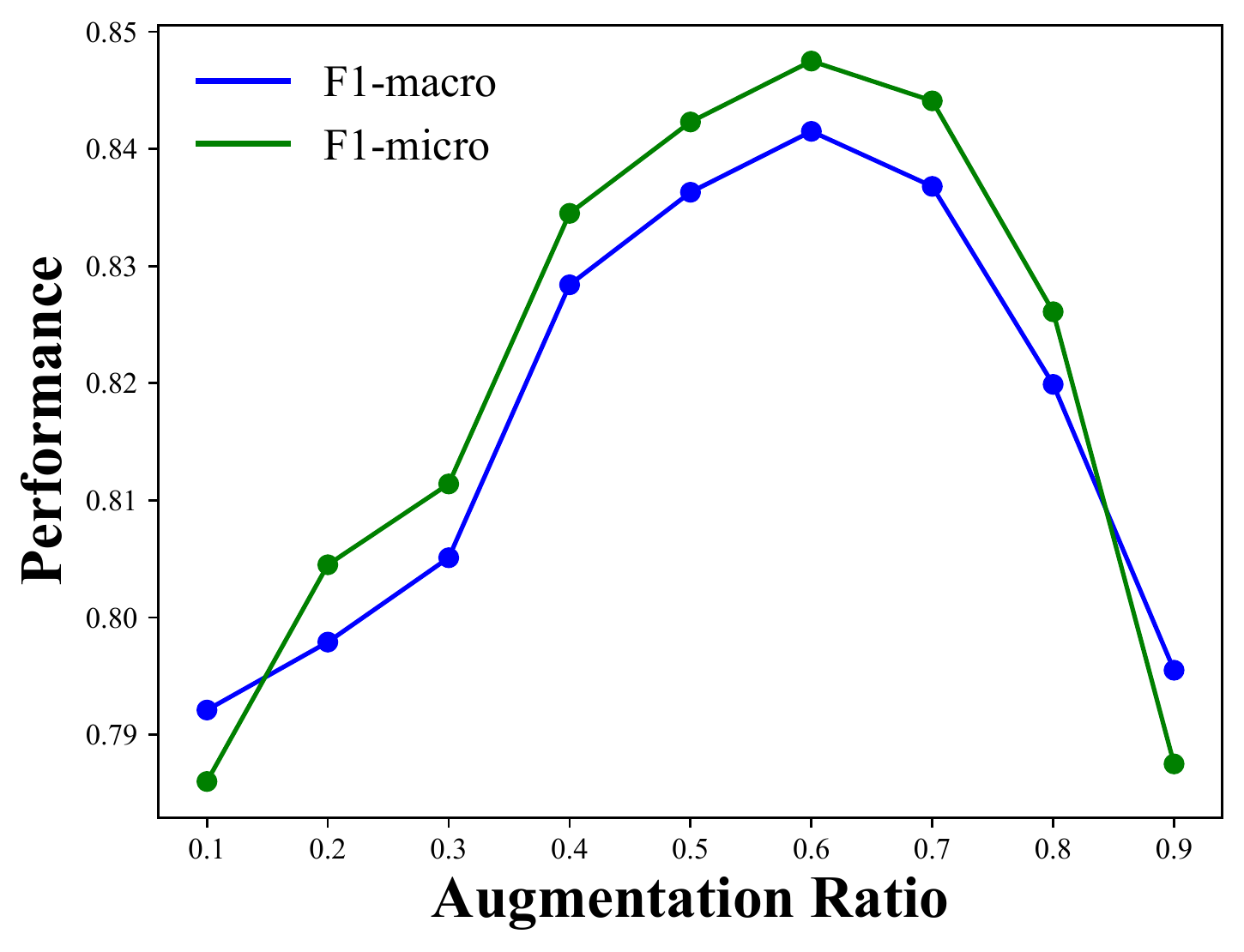}
         \caption{MUTAG}
         \label{fig-MTUAGratio}
     \end{subfigure}
     \begin{subfigure}[b]{0.235\textwidth}
         \centering
         \includegraphics[width=1.00\textwidth]{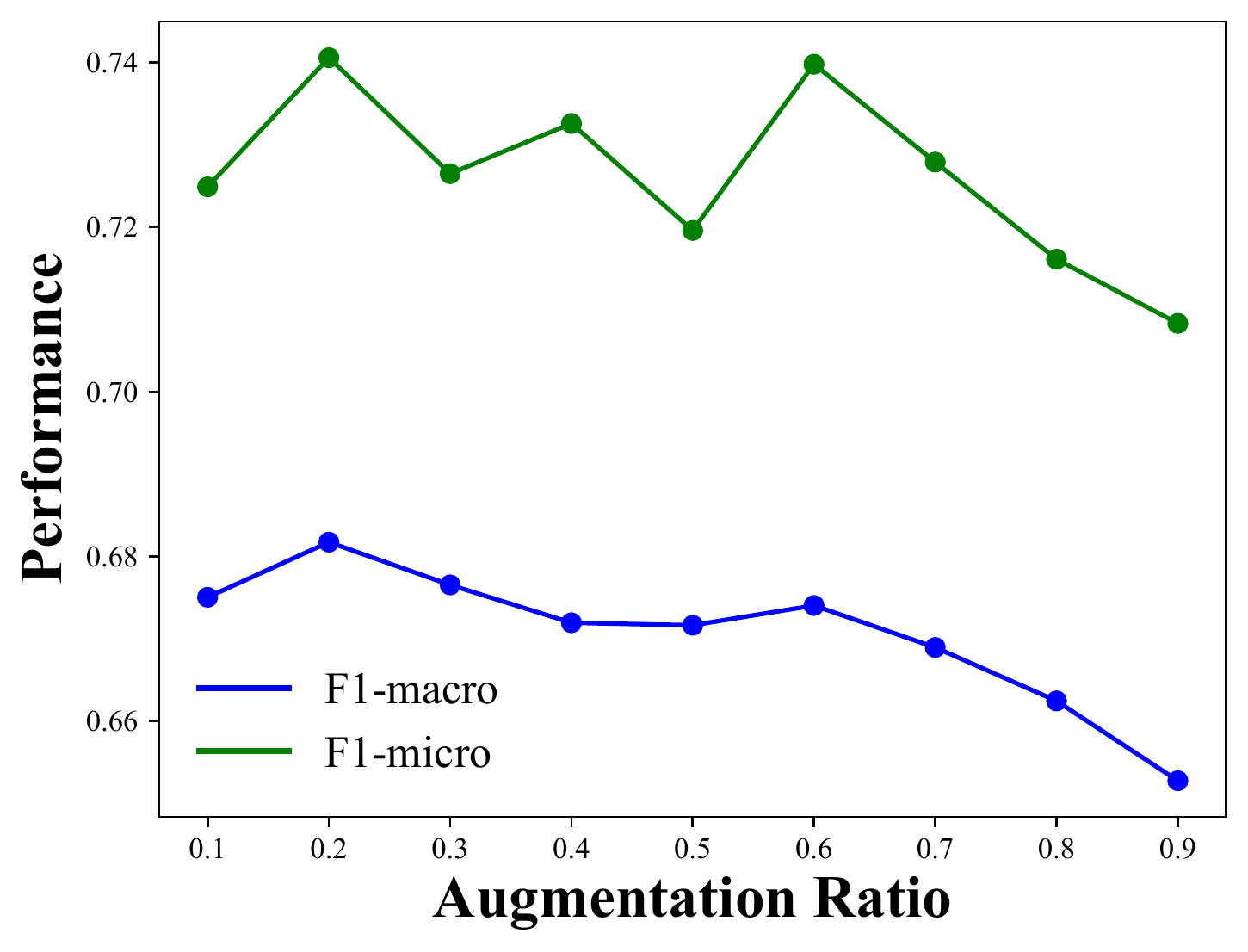}
         \caption{PROTEINS}
         \label{fig-PROTEINSratio}
     \end{subfigure}
     \vskip -1ex
    %  \vspace{-1.5ex}
     \caption{Relationship between augmentation ratio $\delta$ and performance on MUTAG and PROTEINS. The performance first increases and then decreases as augmentation ratio increases. The reported result here is averaged over 20 runs.}
     \vspace{-2ex}
     \label{fig-ratio}
\end{figure}

\vspace{-1.5ex}
\subsubsection{Effect of augmentation ratio}
Then we investigate the effect of augmentation ratio $\delta$ among $\{0.1, 0.2, ..., 0.9\}$ on the performance of G$^2$GNN$_n$ on MUTAG and G$^2$GNN$_e$ on PROTEINS. We see that the F1-macro on both MUTAG and PROTEINS first increase and then decrease. This is because initially increasing augmentation ratio would generate abundant unseen graphs and enhance the model generalibility, which conforms to the advantageous of harder contrastive learning concluded in ~\cite{graphCL}. However, as we further increase the augmentation ratio, the performance decreases because graphs of one class maybe over-augmented, which destroys the latent relationship between graphs and its class or even mismatch graphs with other classes.
% \yy{same: only mention topology, maybe need a small modification here}
% For example, the topology that is typically possessed by graphs in one class maybe augmented towards being similar to the topology in other class, which will misguide the encoder and classifier during training. or even become similar to the topology of other classes and hence the representation we learned may not well-representative of that class.
% \yy{I want to add something for the explanation why increasing the augmentation ratio will reduce the performance. When the ratio increases, the assumption that the generated augmentations for a graph should have a similar prediction distribution no longer holds, which will misguide the training with the classification term and the regularization term. Another thing is that can we include the ratio of zero so that we can verify whether adding augmentation makes a contribution and also it would be clear to see to which point the augmentation will have a lower performance than no augmentation, which I think will provide some insights for interested audience.}

\subsubsection{Efficient Analysis}
Furthermore, we compare the efficiency of each model in Table~\ref{tab-efficiency} where the running time is averaged across 10 times. Without equipping any imbalance-tailored operation, GIN achieves shortest running time. Equipping reweighting as GIN$_\text{rw}$ is faster than equipping upsampling as GIN$_\text{us}$ since upsampling increase the size of the dataset. Our proposed G$^2$GNN and its variants generally have longer running time due to topological augmentation and graph-level propagation. 

\begin{table}[t]
\small
\setlength\tabcolsep{4.5pt}
\caption{Running time (in seconds) of different models.}
\label{tab-efficiency}
\vskip -2ex
\centering
\begin{tabular}{lcccccc}
\hline
\textbf{Dataset} & GIN & GIN$_\text{us}$ & GIN$_\text{rw}$ & G$^2$GNN$_\emptyset$ & G$^2$GNN$_e$& G$^2$GNN$_n$ \\
\hline
MUTAG & 5.2 & 8.9 & 5.6 & 16.8 & 24.4 & 22.6\\
PROTEINS & 24.3 & 40.7 & 25.3 & 111.1 & 155.7 & 153.0\\
\hline
\end{tabular}
\end{table}

\vspace{-1.5ex}
\section{Conclusion}\label{sec-conclusion}
In this paper, we focused on imbalanced graph classification, which widely exists in the real world while rarely explored in the literature. Noticing that unlike the node imbalance problem where we can propagate neighboring nodes' information to obtain extra supervision, graphs are isolated and have no connections with each other. Therefore, we employ a kernel-based Graph of Graph (GoG) construction to establish a kNN graph and devise a two-level propagation to derive extra supervision from neighboring graphs globally. By theoretically proving the feature smoothing is upper bounded by the label smoothing and empirically showing the high homophily on the constructed kNN GoG, we guarantee the derived supervision is beneficial for downstream classification. Moreover, we employ local augmentation and upsampling of minority graphs to enhance the model generalizability in discerning unseen non-training (especially minority) graphs. Experiments on 7 real-world datasets demonstrate the effectiveness of G$^2$GNN in relieving the graph imbalance issue. For future work, we plan to incorporate attention mechanism in the GoG propagation to adaptively aggregate neighboring graphs' information based on their topological similarity and further more work on the imbalance problem in link prediction, especially in recommendation systems~\cite{wang2022collaboration}.

% Moreover, for scenarios where graph topology is non-important, we can simply generalize our proposed GoG module by computing graph similarity based on initial node features rather than topology, which is another potential future direction we prepare to explore.

% \clearpage %just here temporarily to check the reference length to ensure it fits on one page

% \begin{acks}
% \end{acks}
%\balance 
%\vspace{-1ex}
\balance 
\bibliographystyle{ACM-Reference-Format}
\bibliography{references}


\begin{thebibliography}{52}

%%% the Web address.

\ifx \showCODEN    \undefined \def \showCODEN     #1{\unskip}     \fi
\ifx \showDOI      \undefined \def \showDOI       #1{#1}\fi
\ifx \showISBNx    \undefined \def \showISBNx     #1{\unskip}     \fi
\ifx \showISBNxiii \undefined \def \showISBNxiii  #1{\unskip}     \fi
\ifx \showISSN     \undefined \def \showISSN      #1{\unskip}     \fi
\ifx \showLCCN     \undefined \def \showLCCN      #1{\unskip}     \fi
\ifx \shownote     \undefined \def \shownote      #1{#1}          \fi
\ifx \showarticletitle \undefined \def \showarticletitle #1{#1}   \fi
\ifx \showURL      \undefined \def \showURL       {\relax}        \fi
% invisible to TeX
\providecommand\bibfield[2]{#2}
\providecommand\bibinfo[2]{#2}
\providecommand\natexlab[1]{#1}
\providecommand\showeprint[2][]{arXiv:#2}

\bibitem[\protect\citeauthoryear{Berthelot, Carlini, Goodfellow, Papernot,
  Oliver, and Raffel}{Berthelot et~al\mbox{.}}{2019}]%
        {berthelot2019mixmatch}
\bibfield{author}{\bibinfo{person}{David Berthelot}, \bibinfo{person}{Nicholas
  Carlini}, \bibinfo{person}{Ian Goodfellow}, \bibinfo{person}{Nicolas
  Papernot}, \bibinfo{person}{Avital Oliver}, {and} \bibinfo{person}{Colin
  Raffel}.} \bibinfo{year}{2019}\natexlab{}.
\newblock \showarticletitle{Mixmatch: A holistic approach to semi-supervised
  learning}.
\newblock \bibinfo{journal}{\emph{arXiv preprint arXiv:1905.02249}}
  (\bibinfo{year}{2019}).
\newblock


\bibitem[\protect\citeauthoryear{Borgwardt and Kriegel}{Borgwardt and
  Kriegel}{2005}]%
        {borgwardt2005shortest}
\bibfield{author}{\bibinfo{person}{Karsten~M Borgwardt} {and}
  \bibinfo{person}{Hans-Peter Kriegel}.} \bibinfo{year}{2005}\natexlab{}.
\newblock \showarticletitle{Shortest-path kernels on graphs}. In
  \bibinfo{booktitle}{\emph{ICDM}}.
\newblock


\bibitem[\protect\citeauthoryear{Chawla, Bowyer, Hall, and Kegelmeyer}{Chawla
  et~al\mbox{.}}{2002}]%
        {smote}
\bibfield{author}{\bibinfo{person}{Nitesh~V Chawla}, \bibinfo{person}{Kevin~W
  Bowyer}, \bibinfo{person}{Lawrence~O Hall}, {and} \bibinfo{person}{W~Philip
  Kegelmeyer}.} \bibinfo{year}{2002}\natexlab{}.
\newblock \showarticletitle{SMOTE: synthetic minority over-sampling technique}.
\newblock \bibinfo{journal}{\emph{JAIR}}  \bibinfo{volume}{16}
  (\bibinfo{year}{2002}), \bibinfo{pages}{321--357}.
\newblock


\bibitem[\protect\citeauthoryear{Chawla, Japkowicz, and Kotcz}{Chawla
  et~al\mbox{.}}{2004}]%
        {chawla2004special}
\bibfield{author}{\bibinfo{person}{Nitesh~V Chawla}, \bibinfo{person}{Nathalie
  Japkowicz}, {and} \bibinfo{person}{Aleksander Kotcz}.}
  \bibinfo{year}{2004}\natexlab{}.
\newblock \showarticletitle{Special issue on learning from imbalanced data
  sets}.
\newblock \bibinfo{journal}{\emph{SIGKDD Explorations}} \bibinfo{volume}{6},
  \bibinfo{number}{1} (\bibinfo{year}{2004}).
\newblock


\bibitem[\protect\citeauthoryear{Debnath, Lopez~de Compadre, Debnath,
  Shusterman, and Hansch}{Debnath et~al\mbox{.}}{1991}]%
        {debnath1991structure}
\bibfield{author}{\bibinfo{person}{Asim~Kumar Debnath}, \bibinfo{person}{Rosa~L
  Lopez~de Compadre}, \bibinfo{person}{Gargi Debnath}, \bibinfo{person}{Alan~J
  Shusterman}, {and} \bibinfo{person}{Corwin Hansch}.}
  \bibinfo{year}{1991}\natexlab{}.
\newblock \showarticletitle{Structure-activity relationship of mutagenic
  aromatic and heteroaromatic nitro compounds. correlation with molecular
  orbital energies and hydrophobicity}.
\newblock \bibinfo{journal}{\emph{J. Med. Chem.}} \bibinfo{volume}{34},
  \bibinfo{number}{2} (\bibinfo{year}{1991}), \bibinfo{pages}{786--797}.
\newblock


\bibitem[\protect\citeauthoryear{Ding, Wang, Li, Shu, Liu, and Liu}{Ding
  et~al\mbox{.}}{2020}]%
        {gpn}
\bibfield{author}{\bibinfo{person}{Kaize Ding}, \bibinfo{person}{Jianling
  Wang}, \bibinfo{person}{Jundong Li}, \bibinfo{person}{Kai Shu},
  \bibinfo{person}{Chenghao Liu}, {and} \bibinfo{person}{Huan Liu}.}
  \bibinfo{year}{2020}\natexlab{}.
\newblock \showarticletitle{Graph prototypical networks for few-shot learning
  on attributed networks}. In \bibinfo{booktitle}{\emph{CIKM}}.
  \bibinfo{pages}{295--304}.
\newblock


\bibitem[\protect\citeauthoryear{Ding, Xu, Tong, and Liu}{Ding
  et~al\mbox{.}}{2022}]%
        {ding2022data}
\bibfield{author}{\bibinfo{person}{Kaize Ding}, \bibinfo{person}{Zhe Xu},
  \bibinfo{person}{Hanghang Tong}, {and} \bibinfo{person}{Huan Liu}.}
  \bibinfo{year}{2022}\natexlab{}.
\newblock \showarticletitle{Data augmentation for deep graph learning: A
  survey}.
\newblock \bibinfo{journal}{\emph{arXiv preprint arXiv:2202.08235}}
  (\bibinfo{year}{2022}).
\newblock


\bibitem[\protect\citeauthoryear{Errica, Podda, Bacciu, and Micheli}{Errica
  et~al\mbox{.}}{2020}]%
        {fairgraph}
\bibfield{author}{\bibinfo{person}{Federico Errica}, \bibinfo{person}{Marco
  Podda}, \bibinfo{person}{Davide Bacciu}, {and} \bibinfo{person}{Alessio
  Micheli}.} \bibinfo{year}{2020}\natexlab{}.
\newblock \showarticletitle{A Fair Comparison of Graph Neural Networks for
  Graph Classification}. In \bibinfo{booktitle}{\emph{ICLR}}.
\newblock


\bibitem[\protect\citeauthoryear{Feng, Gangal, Wei, Chandar, Vosoughi,
  Mitamura, and Hovy}{Feng et~al\mbox{.}}{2021}]%
        {feng2021survey}
\bibfield{author}{\bibinfo{person}{Steven~Y Feng}, \bibinfo{person}{Varun
  Gangal}, \bibinfo{person}{Jason Wei}, \bibinfo{person}{Sarath Chandar},
  \bibinfo{person}{Soroush Vosoughi}, \bibinfo{person}{Teruko Mitamura}, {and}
  \bibinfo{person}{Eduard Hovy}.} \bibinfo{year}{2021}\natexlab{}.
\newblock \showarticletitle{A survey of data augmentation approaches for nlp}.
\newblock \bibinfo{journal}{\emph{arXiv:2105.03075}} (\bibinfo{year}{2021}).
\newblock


\bibitem[\protect\citeauthoryear{Feng, Zhang, Dong, Han, Luan, Xu, Yang,
  Kharlamov, and Tang}{Feng et~al\mbox{.}}{2020}]%
        {GRAND}
\bibfield{author}{\bibinfo{person}{Wenzheng Feng}, \bibinfo{person}{Jie Zhang},
  \bibinfo{person}{Yuxiao Dong}, \bibinfo{person}{Yu Han},
  \bibinfo{person}{Huanbo Luan}, \bibinfo{person}{Qian Xu},
  \bibinfo{person}{Qiang Yang}, \bibinfo{person}{Evgeny Kharlamov}, {and}
  \bibinfo{person}{Jie Tang}.} \bibinfo{year}{2020}\natexlab{}.
\newblock \showarticletitle{Graph Random Neural Network for Semi-Supervised
  Learning on Graphs}.
\newblock \bibinfo{journal}{\emph{arXiv:2005.11079}} (\bibinfo{year}{2020}).
\newblock


\bibitem[\protect\citeauthoryear{Fey and Lenssen}{Fey and Lenssen}{2019}]%
        {fey2019fast}
\bibfield{author}{\bibinfo{person}{Matthias Fey} {and}
  \bibinfo{person}{Jan~Eric Lenssen}.} \bibinfo{year}{2019}\natexlab{}.
\newblock \showarticletitle{Fast graph representation learning with PyTorch
  Geometric}.
\newblock \bibinfo{journal}{\emph{arXiv preprint arXiv:1903.02428}}
  (\bibinfo{year}{2019}).
\newblock


\bibitem[\protect\citeauthoryear{Grandvalet and Bengio}{Grandvalet and
  Bengio}{2004}]%
        {grandvalet2004semi}
\bibfield{author}{\bibinfo{person}{Yves Grandvalet} {and}
  \bibinfo{person}{Yoshua Bengio}.} \bibinfo{year}{2004}\natexlab{}.
\newblock \showarticletitle{Semi-supervised learning by entropy minimization}.
\newblock \bibinfo{journal}{\emph{Advances in neural information processing
  systems}}  \bibinfo{volume}{17} (\bibinfo{year}{2004}).
\newblock


\bibitem[\protect\citeauthoryear{Gu, Jiang, Guzzi, and Milenkovic}{Gu
  et~al\mbox{.}}{2021}]%
        {gu2021modeling}
\bibfield{author}{\bibinfo{person}{Shawn Gu}, \bibinfo{person}{Meng Jiang},
  \bibinfo{person}{Pietro~Hiram Guzzi}, {and} \bibinfo{person}{Tijana
  Milenkovic}.} \bibinfo{year}{2021}\natexlab{}.
\newblock \showarticletitle{Modeling multi-scale data via a network of
  networks}.
\newblock \bibinfo{journal}{\emph{arXiv:2105.12226}} (\bibinfo{year}{2021}).
\newblock


\bibitem[\protect\citeauthoryear{Idakwo, Thangapandian, Luttrell, Li, Wang,
  Zhou, Hong, Yang, Zhang, and Gong}{Idakwo et~al\mbox{.}}{2020}]%
        {idakwo2020structure}
\bibfield{author}{\bibinfo{person}{Gabriel Idakwo}, \bibinfo{person}{Sundar
  Thangapandian}, \bibinfo{person}{Joseph Luttrell}, \bibinfo{person}{Yan Li},
  \bibinfo{person}{Nan Wang}, \bibinfo{person}{Zhaoxian Zhou},
  \bibinfo{person}{Huixiao Hong}, \bibinfo{person}{Bei Yang},
  \bibinfo{person}{Chaoyang Zhang}, {and} \bibinfo{person}{Ping Gong}.}
  \bibinfo{year}{2020}\natexlab{}.
\newblock \showarticletitle{Structure--activity relationship-based chemical
  classification of highly imbalanced Tox21 datasets}.
\newblock \bibinfo{journal}{\emph{Journal of cheminformatics}}
  \bibinfo{volume}{12}, \bibinfo{number}{1} (\bibinfo{year}{2020}),
  \bibinfo{pages}{1--19}.
\newblock


\bibitem[\protect\citeauthoryear{Johnson and Khoshgoftaar}{Johnson and
  Khoshgoftaar}{2019}]%
        {johnson2019survey}
\bibfield{author}{\bibinfo{person}{Justin~M Johnson} {and}
  \bibinfo{person}{Taghi~M Khoshgoftaar}.} \bibinfo{year}{2019}\natexlab{}.
\newblock \showarticletitle{Survey on deep learning with class imbalance}.
\newblock \bibinfo{journal}{\emph{Journal of Big Data}} \bibinfo{volume}{6},
  \bibinfo{number}{1} (\bibinfo{year}{2019}), \bibinfo{pages}{1--54}.
\newblock


\bibitem[\protect\citeauthoryear{Kilhamn}{Kilhamn}{2015}]%
        {kilhamn2015fast}
\bibfield{author}{\bibinfo{person}{Jonatan Kilhamn}.}
  \bibinfo{year}{2015}\natexlab{}.
\newblock \emph{\bibinfo{title}{Fast shortest-path kernel computations using
  aproximate methods}}.
\newblock \bibinfo{thesistype}{Master's\ thesis}.
\newblock


\bibitem[\protect\citeauthoryear{Kubat, Matwin, et~al\mbox{.}}{Kubat
  et~al\mbox{.}}{1997}]%
        {kubat1997addressing}
\bibfield{author}{\bibinfo{person}{Miroslav Kubat}, \bibinfo{person}{Stan
  Matwin}, {et~al\mbox{.}}} \bibinfo{year}{1997}\natexlab{}.
\newblock \showarticletitle{Addressing the curse of imbalanced training sets:
  one-sided selection}. In \bibinfo{booktitle}{\emph{Icml}},
  Vol.~\bibinfo{volume}{97}. Citeseer, \bibinfo{pages}{179}.
\newblock


\bibitem[\protect\citeauthoryear{Leevy, Khoshgoftaar, Bauder, and Seliya}{Leevy
  et~al\mbox{.}}{2018}]%
        {leevy2018survey}
\bibfield{author}{\bibinfo{person}{Joffrey~L Leevy}, \bibinfo{person}{Taghi~M
  Khoshgoftaar}, \bibinfo{person}{Richard~A Bauder}, {and}
  \bibinfo{person}{Naeem Seliya}.} \bibinfo{year}{2018}\natexlab{}.
\newblock \showarticletitle{A survey on addressing high-class imbalance in big
  data}.
\newblock \bibinfo{journal}{\emph{Journal of Big Data}} \bibinfo{volume}{5},
  \bibinfo{number}{1} (\bibinfo{year}{2018}), \bibinfo{pages}{1--30}.
\newblock


\bibitem[\protect\citeauthoryear{Li, Han, and Wu}{Li et~al\mbox{.}}{2018}]%
        {li2018deeper}
\bibfield{author}{\bibinfo{person}{Qimai Li}, \bibinfo{person}{Zhichao Han},
  {and} \bibinfo{person}{Xiao-Ming Wu}.} \bibinfo{year}{2018}\natexlab{}.
\newblock \showarticletitle{Deeper insights into graph convolutional networks
  for semi-supervised learning}. In \bibinfo{booktitle}{\emph{AAAI}}.
\newblock


\bibitem[\protect\citeauthoryear{Liu, Gao, and Ji}{Liu et~al\mbox{.}}{2020}]%
        {liu2020towards}
\bibfield{author}{\bibinfo{person}{Meng Liu}, \bibinfo{person}{Hongyang Gao},
  {and} \bibinfo{person}{Shuiwang Ji}.} \bibinfo{year}{2020}\natexlab{}.
\newblock \showarticletitle{Towards deeper graph neural networks}. In
  \bibinfo{booktitle}{\emph{Proceedings of the 26th ACM SIGKDD international
  conference on knowledge discovery \& data mining}}.
  \bibinfo{pages}{338--348}.
\newblock


\bibitem[\protect\citeauthoryear{Liu, Wang, Vu, Moretti, Bodenheimer, Meiler,
  and Derr}{Liu et~al\mbox{.}}{2022}]%
        {liu2022interpretable}
\bibfield{author}{\bibinfo{person}{Yunchao Liu}, \bibinfo{person}{Yu Wang},
  \bibinfo{person}{Oanh~T Vu}, \bibinfo{person}{Rocco Moretti},
  \bibinfo{person}{Bobby Bodenheimer}, \bibinfo{person}{Jens Meiler}, {and}
  \bibinfo{person}{Tyler Derr}.} \bibinfo{year}{2022}\natexlab{}.
\newblock \showarticletitle{Interpretable Chirality-Aware Graph Neural Network
  for Quantitative Structure Activity Relationship Modeling in Drug Discovery}.
\newblock \bibinfo{journal}{\emph{bioRxiv}} (\bibinfo{year}{2022}).
\newblock


\bibitem[\protect\citeauthoryear{Ni, Tong, Fan, and Zhang}{Ni
  et~al\mbox{.}}{2014}]%
        {non}
\bibfield{author}{\bibinfo{person}{Jingchao Ni}, \bibinfo{person}{Hanghang
  Tong}, \bibinfo{person}{Wei Fan}, {and} \bibinfo{person}{Xiang Zhang}.}
  \bibinfo{year}{2014}\natexlab{}.
\newblock \showarticletitle{Inside the atoms: ranking on a network of
  networks}. In \bibinfo{booktitle}{\emph{KDD}}. \bibinfo{pages}{1356--1365}.
\newblock


\bibitem[\protect\citeauthoryear{Pan and Zhu}{Pan and Zhu}{2013}]%
        {pan2013graph}
\bibfield{author}{\bibinfo{person}{Shirui Pan} {and} \bibinfo{person}{Xingquan
  Zhu}.} \bibinfo{year}{2013}\natexlab{}.
\newblock \showarticletitle{Graph classification with imbalanced class
  distributions and noise}. In \bibinfo{booktitle}{\emph{IJCAI}}.
\newblock


\bibitem[\protect\citeauthoryear{Qu, Zhu, Zheng, Shi, and Yin}{Qu
  et~al\mbox{.}}{2021}]%
        {qu2021imgagn}
\bibfield{author}{\bibinfo{person}{Liang Qu}, \bibinfo{person}{Huaisheng Zhu},
  \bibinfo{person}{Ruiqi Zheng}, \bibinfo{person}{Yuhui Shi}, {and}
  \bibinfo{person}{Hongzhi Yin}.} \bibinfo{year}{2021}\natexlab{}.
\newblock \showarticletitle{ImGAGN: Imbalanced Network Embedding via Generative
  Adversarial Graph Networks}.
\newblock \bibinfo{journal}{\emph{arXiv:2106.02817}} (\bibinfo{year}{2021}).
\newblock


\bibitem[\protect\citeauthoryear{Rozemberczki, Hoyt, Gogleva, Grabowski, Karis,
  Lamov, Nikolov, Nilsson, Ughetto, Wang, et~al\mbox{.}}{Rozemberczki
  et~al\mbox{.}}{2022}]%
        {rozemberczki2022chemicalx}
\bibfield{author}{\bibinfo{person}{Benedek Rozemberczki},
  \bibinfo{person}{Charles~Tapley Hoyt}, \bibinfo{person}{Anna Gogleva},
  \bibinfo{person}{Piotr Grabowski}, \bibinfo{person}{Klas Karis},
  \bibinfo{person}{Andrej Lamov}, \bibinfo{person}{Andriy Nikolov},
  \bibinfo{person}{Sebastian Nilsson}, \bibinfo{person}{Michael Ughetto},
  \bibinfo{person}{Yu Wang}, {et~al\mbox{.}}} \bibinfo{year}{2022}\natexlab{}.
\newblock \showarticletitle{ChemicalX: A Deep Learning Library for Drug Pair
  Scoring}. In \bibinfo{booktitle}{\emph{Proceedings of the 28th ACM SIGKDD
  Conference on Knowledge Discovery and Data Mining}}.
  \bibinfo{pages}{3819--3828}.
\newblock


\bibitem[\protect\citeauthoryear{Shi, Tang, Zhu, Wilson, and Liu}{Shi
  et~al\mbox{.}}{2020}]%
        {DRGCN}
\bibfield{author}{\bibinfo{person}{Min Shi}, \bibinfo{person}{Yufei Tang},
  \bibinfo{person}{Xingquan Zhu}, \bibinfo{person}{David Wilson}, {and}
  \bibinfo{person}{Jianxun Liu}.} \bibinfo{year}{2020}\natexlab{}.
\newblock \showarticletitle{Multi-class imbalanced graph convolutional network
  learning}. In \bibinfo{booktitle}{\emph{IJCAI}}.
\newblock


\bibitem[\protect\citeauthoryear{Shorten and Khoshgoftaar}{Shorten and
  Khoshgoftaar}{2019}]%
        {shorten2019survey}
\bibfield{author}{\bibinfo{person}{Connor Shorten} {and}
  \bibinfo{person}{Taghi~M Khoshgoftaar}.} \bibinfo{year}{2019}\natexlab{}.
\newblock \showarticletitle{A survey on image data augmentation for deep
  learning}.
\newblock \bibinfo{journal}{\emph{Journal of Big Data}} \bibinfo{volume}{6},
  \bibinfo{number}{1} (\bibinfo{year}{2019}), \bibinfo{pages}{1--48}.
\newblock


\bibitem[\protect\citeauthoryear{Song, Yang, Xu, and King}{Song
  et~al\mbox{.}}{2021}]%
        {song2021graph}
\bibfield{author}{\bibinfo{person}{Zixing Song}, \bibinfo{person}{Xiangli
  Yang}, \bibinfo{person}{Zenglin Xu}, {and} \bibinfo{person}{Irwin King}.}
  \bibinfo{year}{2021}\natexlab{}.
\newblock \showarticletitle{Graph-based Semi-supervised Learning: A
  Comprehensive Review}.
\newblock \bibinfo{journal}{\emph{arXiv:2102.13303}} (\bibinfo{year}{2021}).
\newblock


\bibitem[\protect\citeauthoryear{Sun, Hoffmann, Verma, and Tang}{Sun
  et~al\mbox{.}}{[n.d.]}]%
        {Infograph}
\bibfield{author}{\bibinfo{person}{Fan{-}Yun Sun}, \bibinfo{person}{Jordan
  Hoffmann}, \bibinfo{person}{Vikas Verma}, {and} \bibinfo{person}{Jian Tang}.}
  \bibinfo{year}{[n.d.]}\natexlab{}.
\newblock \showarticletitle{InfoGraph: Unsupervised and Semi-supervised
  Graph-Level Representation Learning via Mutual Information Maximization}. In
  \bibinfo{booktitle}{\emph{ICLR 2020}}.
\newblock


\bibitem[\protect\citeauthoryear{Sutherland, O'brien, and Weaver}{Sutherland
  et~al\mbox{.}}{2003}]%
        {sutherland2003spline}
\bibfield{author}{\bibinfo{person}{Jeffrey~J Sutherland},
  \bibinfo{person}{Lee~A O'brien}, {and} \bibinfo{person}{Donald~F Weaver}.}
  \bibinfo{year}{2003}\natexlab{}.
\newblock \showarticletitle{Spline-fitting with a genetic algorithm: A method
  for developing classification structure- activity relationships}.
\newblock \bibinfo{journal}{\emph{J Chem Inform Comput Sci}}
  \bibinfo{volume}{43}, \bibinfo{number}{6} (\bibinfo{year}{2003}).
\newblock


\bibitem[\protect\citeauthoryear{Thai-Nghe, Gantner, and
  Schmidt-Thieme}{Thai-Nghe et~al\mbox{.}}{2010}]%
        {thai2010cost}
\bibfield{author}{\bibinfo{person}{Nguyen Thai-Nghe}, \bibinfo{person}{Zeno
  Gantner}, {and} \bibinfo{person}{Lars Schmidt-Thieme}.}
  \bibinfo{year}{2010}\natexlab{}.
\newblock \showarticletitle{Cost-sensitive learning methods for imbalanced
  data}. In \bibinfo{booktitle}{\emph{IJCNN}}. IEEE.
\newblock


\bibitem[\protect\citeauthoryear{Toivonen, Srinivasan, King, Kramer, and
  Helma}{Toivonen et~al\mbox{.}}{2003}]%
        {toivonen2003statistical}
\bibfield{author}{\bibinfo{person}{Hannu Toivonen}, \bibinfo{person}{Ashwin
  Srinivasan}, \bibinfo{person}{Ross~D King}, \bibinfo{person}{Stefan Kramer},
  {and} \bibinfo{person}{Christoph Helma}.} \bibinfo{year}{2003}\natexlab{}.
\newblock \showarticletitle{Statistical evaluation of the predictive toxicology
  challenge 2000--2001}.
\newblock \bibinfo{journal}{\emph{Bioinformatics}} \bibinfo{volume}{19},
  \bibinfo{number}{10} (\bibinfo{year}{2003}), \bibinfo{pages}{1183--1193}.
\newblock


\bibitem[\protect\citeauthoryear{Van~Hulse, Khoshgoftaar, and
  Napolitano}{Van~Hulse et~al\mbox{.}}{2007}]%
        {van2007experimental}
\bibfield{author}{\bibinfo{person}{Jason Van~Hulse}, \bibinfo{person}{Taghi~M
  Khoshgoftaar}, {and} \bibinfo{person}{Amri Napolitano}.}
  \bibinfo{year}{2007}\natexlab{}.
\newblock \showarticletitle{Experimental perspectives on learning from
  imbalanced data}. In \bibinfo{booktitle}{\emph{ICML}}.
  \bibinfo{pages}{935--942}.
\newblock


\bibitem[\protect\citeauthoryear{Vishwanathan, Schraudolph, Kondor, and
  Borgwardt}{Vishwanathan et~al\mbox{.}}{2010}]%
        {vishwanathan2010graph}
\bibfield{author}{\bibinfo{person}{S~Vichy~N Vishwanathan},
  \bibinfo{person}{Nicol~N Schraudolph}, \bibinfo{person}{Risi Kondor}, {and}
  \bibinfo{person}{Karsten~M Borgwardt}.} \bibinfo{year}{2010}\natexlab{}.
\newblock \showarticletitle{Graph kernels}.
\newblock \bibinfo{journal}{\emph{Journal of Machine Learning Research}}
  \bibinfo{volume}{11} (\bibinfo{year}{2010}), \bibinfo{pages}{1201--1242}.
\newblock


\bibitem[\protect\citeauthoryear{Wale, Watson, and Karypis}{Wale
  et~al\mbox{.}}{2008}]%
        {wale2008comparison}
\bibfield{author}{\bibinfo{person}{Nikil Wale}, \bibinfo{person}{Ian~A Watson},
  {and} \bibinfo{person}{George Karypis}.} \bibinfo{year}{2008}\natexlab{}.
\newblock \showarticletitle{Comparison of descriptor spaces for chemical
  compound retrieval and classification}.
\newblock \bibinfo{journal}{\emph{KAIS}} \bibinfo{volume}{14},
  \bibinfo{number}{3} (\bibinfo{year}{2008}), \bibinfo{pages}{347--375}.
\newblock


\bibitem[\protect\citeauthoryear{Wang and Leskovec}{Wang and Leskovec}{2020}]%
        {wang2020unifying}
\bibfield{author}{\bibinfo{person}{Hongwei Wang} {and} \bibinfo{person}{Jure
  Leskovec}.} \bibinfo{year}{2020}\natexlab{}.
\newblock \showarticletitle{Unifying graph convolutional neural networks and
  label propagation}.
\newblock \bibinfo{journal}{\emph{arXiv preprint arXiv:2002.06755}}
  (\bibinfo{year}{2020}).
\newblock


\bibitem[\protect\citeauthoryear{Wang, Lian, Zhang, Qin, and Lin}{Wang
  et~al\mbox{.}}{2020a}]%
        {wang2020gognn}
\bibfield{author}{\bibinfo{person}{Hanchen Wang}, \bibinfo{person}{Defu Lian},
  \bibinfo{person}{Ying Zhang}, \bibinfo{person}{Lu Qin}, {and}
  \bibinfo{person}{Xuemin Lin}.} \bibinfo{year}{2020}\natexlab{a}.
\newblock \showarticletitle{Gognn: Graph of graphs neural network for
  predicting structured entity interactions}.
\newblock \bibinfo{journal}{\emph{arXiv:2005.05537}} (\bibinfo{year}{2020}).
\newblock


\bibitem[\protect\citeauthoryear{Wang, Aggarwal, and Derr}{Wang
  et~al\mbox{.}}{2021}]%
        {wang2021distance}
\bibfield{author}{\bibinfo{person}{Yu Wang}, \bibinfo{person}{Charu Aggarwal},
  {and} \bibinfo{person}{Tyler Derr}.} \bibinfo{year}{2021}\natexlab{}.
\newblock \showarticletitle{Distance-wise Prototypical Graph Neural Network in
  Node Imbalance Classification}.
\newblock \bibinfo{journal}{\emph{arXiv preprint arXiv:2110.12035}}
  (\bibinfo{year}{2021}).
\newblock


\bibitem[\protect\citeauthoryear{Wang, Jin, and Derr}{Wang
  et~al\mbox{.}}{2022a}]%
        {wang2022graph}
\bibfield{author}{\bibinfo{person}{Yu Wang}, \bibinfo{person}{Wei Jin}, {and}
  \bibinfo{person}{Tyler Derr}.} \bibinfo{year}{2022}\natexlab{a}.
\newblock \showarticletitle{Graph neural networks: Self-supervised learning}.
\newblock In \bibinfo{booktitle}{\emph{Graph Neural Networks: Foundations,
  Frontiers, and Applications}}. \bibinfo{publisher}{Springer},
  \bibinfo{pages}{391--420}.
\newblock


\bibitem[\protect\citeauthoryear{Wang, Zhao, Zhang, and Derr}{Wang
  et~al\mbox{.}}{2022b}]%
        {wang2022collaboration}
\bibfield{author}{\bibinfo{person}{Yu Wang}, \bibinfo{person}{Yuying Zhao},
  \bibinfo{person}{Yi Zhang}, {and} \bibinfo{person}{Tyler Derr}.}
  \bibinfo{year}{2022}\natexlab{b}.
\newblock \showarticletitle{Collaboration-Aware Graph Convolutional Networks
  for Recommendation Systems}.
\newblock \bibinfo{journal}{\emph{arXiv preprint arXiv:2207.06221}}
  (\bibinfo{year}{2022}).
\newblock


\bibitem[\protect\citeauthoryear{Wang, Ye, Wang, Cui, and Yu}{Wang
  et~al\mbox{.}}{2020b}]%
        {RECT}
\bibfield{author}{\bibinfo{person}{Zheng Wang}, \bibinfo{person}{Xiaojun Ye},
  \bibinfo{person}{Chaokun Wang}, \bibinfo{person}{Jian Cui}, {and}
  \bibinfo{person}{Philip Yu}.} \bibinfo{year}{2020}\natexlab{b}.
\newblock \showarticletitle{Network embedding with completely-imbalanced
  labels}.
\newblock \bibinfo{journal}{\emph{IEEE TKDE}} (\bibinfo{year}{2020}).
\newblock


\bibitem[\protect\citeauthoryear{Xu, Hu, Leskovec, and Jegelka}{Xu
  et~al\mbox{.}}{2019}]%
        {powerfulgnn}
\bibfield{author}{\bibinfo{person}{Keyulu Xu}, \bibinfo{person}{Weihua Hu},
  \bibinfo{person}{Jure Leskovec}, {and} \bibinfo{person}{Stefanie Jegelka}.}
  \bibinfo{year}{2019}\natexlab{}.
\newblock \showarticletitle{How Powerful are Graph Neural Networks?}. In
  \bibinfo{booktitle}{\emph{ICLR}}.
\newblock


\bibitem[\protect\citeauthoryear{Yanardag and Vishwanathan}{Yanardag and
  Vishwanathan}{2015}]%
        {yanardag2015deep}
\bibfield{author}{\bibinfo{person}{Pinar Yanardag} {and} \bibinfo{person}{SVN
  Vishwanathan}.} \bibinfo{year}{2015}\natexlab{}.
\newblock \showarticletitle{Deep graph kernels}. In
  \bibinfo{booktitle}{\emph{KDD}}. \bibinfo{pages}{1365--1374}.
\newblock


\bibitem[\protect\citeauthoryear{You, Chen, Sui, Chen, Wang, and Shen}{You
  et~al\mbox{.}}{2020}]%
        {graphCL}
\bibfield{author}{\bibinfo{person}{Yuning You}, \bibinfo{person}{Tianlong
  Chen}, \bibinfo{person}{Yongduo Sui}, \bibinfo{person}{Ting Chen},
  \bibinfo{person}{Zhangyang Wang}, {and} \bibinfo{person}{Yang Shen}.}
  \bibinfo{year}{2020}\natexlab{}.
\newblock \showarticletitle{Graph Contrastive Learning with Augmentations}. In
  \bibinfo{booktitle}{\emph{NeurIPS}}.
\newblock


\bibitem[\protect\citeauthoryear{Yu, Wu, Bichler, Aros-Vera, and Gao}{Yu
  et~al\mbox{.}}{2022}]%
        {yu2022reconstructing}
\bibfield{author}{\bibinfo{person}{Jin-Zhu Yu}, \bibinfo{person}{Mincheng Wu},
  \bibinfo{person}{Gisela Bichler}, \bibinfo{person}{Felipe Aros-Vera}, {and}
  \bibinfo{person}{Jianxi Gao}.} \bibinfo{year}{2022}\natexlab{}.
\newblock \showarticletitle{Reconstructing Sparse Illicit Supply Networks: A
  Case Study of Multiplex Drug Trafficking Networks}.
\newblock \bibinfo{journal}{\emph{arXiv preprint arXiv:2208.01739}}
  (\bibinfo{year}{2022}).
\newblock


\bibitem[\protect\citeauthoryear{Yuan and Ma}{Yuan and Ma}{2012}]%
        {yuan2012sampling+}
\bibfield{author}{\bibinfo{person}{Bo Yuan} {and} \bibinfo{person}{Xiaoli Ma}.}
  \bibinfo{year}{2012}\natexlab{}.
\newblock \showarticletitle{Sampling+ reweighting: Boosting the performance of
  AdaBoost on imbalanced datasets}. In \bibinfo{booktitle}{\emph{The 2012
  international joint conference on neural networks (IJCNN)}}. IEEE,
  \bibinfo{pages}{1--6}.
\newblock


\bibitem[\protect\citeauthoryear{Zhang, Cisse, Dauphin, and Lopez-Paz}{Zhang
  et~al\mbox{.}}{2017}]%
        {mix}
\bibfield{author}{\bibinfo{person}{Hongyi Zhang}, \bibinfo{person}{Moustapha
  Cisse}, \bibinfo{person}{Yann~N Dauphin}, {and} \bibinfo{person}{David
  Lopez-Paz}.} \bibinfo{year}{2017}\natexlab{}.
\newblock \showarticletitle{mixup: Beyond empirical risk minimization}.
\newblock \bibinfo{journal}{\emph{arXiv preprint arXiv:1710.09412}}
  (\bibinfo{year}{2017}).
\newblock


\bibitem[\protect\citeauthoryear{Zhao, Jiang, Shah, and Jiang}{Zhao
  et~al\mbox{.}}{2021a}]%
        {zhao2021synergistic}
\bibfield{author}{\bibinfo{person}{Tong Zhao}, \bibinfo{person}{Tianwen Jiang},
  \bibinfo{person}{Neil Shah}, {and} \bibinfo{person}{Meng Jiang}.}
  \bibinfo{year}{2021}\natexlab{a}.
\newblock \showarticletitle{A Synergistic Approach for Graph Anomaly Detection
  With Pattern Mining and Feature Learning}.
\newblock \bibinfo{journal}{\emph{IEEE TNNLS}} (\bibinfo{year}{2021}).
\newblock


\bibitem[\protect\citeauthoryear{Zhao, Liu, G{\"u}nnemann, and Jiang}{Zhao
  et~al\mbox{.}}{2022}]%
        {zhao2022graph}
\bibfield{author}{\bibinfo{person}{Tong Zhao}, \bibinfo{person}{Gang Liu},
  \bibinfo{person}{Stephan G{\"u}nnemann}, {and} \bibinfo{person}{Meng Jiang}.}
  \bibinfo{year}{2022}\natexlab{}.
\newblock \showarticletitle{Graph Data Augmentation for Graph Machine Learning:
  A Survey}.
\newblock \bibinfo{journal}{\emph{arXiv preprint arXiv:2202.08871}}
  (\bibinfo{year}{2022}).
\newblock


\bibitem[\protect\citeauthoryear{Zhao, Liu, Neves, Woodford, Jiang, and
  Shah}{Zhao et~al\mbox{.}}{2021b}]%
        {zhao2021data}
\bibfield{author}{\bibinfo{person}{Tong Zhao}, \bibinfo{person}{Yozen Liu},
  \bibinfo{person}{Leonardo Neves}, \bibinfo{person}{Oliver Woodford},
  \bibinfo{person}{Meng Jiang}, {and} \bibinfo{person}{Neil Shah}.}
  \bibinfo{year}{2021}\natexlab{b}.
\newblock \showarticletitle{Data augmentation for graph neural networks}. In
  \bibinfo{booktitle}{\emph{Proceedings of the AAAI Conference on Artificial
  Intelligence}}, Vol.~\bibinfo{volume}{35}. \bibinfo{pages}{11015--11023}.
\newblock


\bibitem[\protect\citeauthoryear{Zhao, Zhang, and Wang}{Zhao
  et~al\mbox{.}}{2021c}]%
        {graphsmote}
\bibfield{author}{\bibinfo{person}{Tianxiang Zhao}, \bibinfo{person}{Xiang
  Zhang}, {and} \bibinfo{person}{Suhang Wang}.}
  \bibinfo{year}{2021}\natexlab{c}.
\newblock \showarticletitle{GraphSMOTE: Imbalanced Node Classification on
  Graphs with Graph Neural Networks}. In \bibinfo{booktitle}{\emph{WSDM}}.
\newblock


\bibitem[\protect\citeauthoryear{Zheng, Lei, Ai, Chen, Deng, and Yang}{Zheng
  et~al\mbox{.}}{2021}]%
        {zheng2021deep}
\bibfield{author}{\bibinfo{person}{Shuangjia Zheng}, \bibinfo{person}{Zengrong
  Lei}, \bibinfo{person}{Haitao Ai}, \bibinfo{person}{Hongming Chen},
  \bibinfo{person}{Daiguo Deng}, {and} \bibinfo{person}{Yuedong Yang}.}
  \bibinfo{year}{2021}\natexlab{}.
\newblock \showarticletitle{Deep scaffold hopping with multimodal transformer
  neural networks}.
\newblock \bibinfo{journal}{\emph{Journal of cheminformatics}}
  \bibinfo{volume}{13}, \bibinfo{number}{1} (\bibinfo{year}{2021}),
  \bibinfo{pages}{1--15}.
\newblock


\end{thebibliography}

% \appendix








% \clearpage
%\newpage
% \pagebreak
% \appendix
% \input{Appendix}

\end{document}